\documentclass[a4paper, 11pt]{article}
\usepackage{multicol}
\usepackage{mwe,lipsum}
\usepackage{amsmath,amsfonts}
\usepackage{algorithmic}
\usepackage{algorithm}
\usepackage{array}
\usepackage[caption=false,font=normalsize,labelfont=sf,textfont=sf]{subfig}
\usepackage{textcomp}
\usepackage{stfloats}
\usepackage{url}
\usepackage{verbatim}
\usepackage{graphicx}
\usepackage{cite}
\usepackage{epstopdf}
\usepackage{epsf}
\usepackage{graphicx}
\usepackage{comment}
\usepackage{amsmath}
\usepackage{pdflscape}
\usepackage{amssymb}
\usepackage{amsmath}
\usepackage{bm}
\usepackage{booktabs}
\usepackage{multicol}
\usepackage{algorithmic}
\usepackage{algorithm}
\usepackage[dvips]{epsfig}
\usepackage[usenames, dvipsnames, table]{xcolor}
\usepackage{tikz}
\usetikzlibrary{calc,shapes,arrows,plotmarks}
\usepackage{layout}
\usepackage{multirow}
\usepackage{color}
\usepackage{afterpage}
\usepackage{balance}
\usepackage[ansinew]{inputenc}
\usepackage{array}
\newtheorem{definition}{Definition}

\begin{document}

\title{Metaheuristics for (Variable-Size) Mixed Optimization Problems: \\
	A Unified Taxonomy and Survey}

\author{Prof. El-Ghazali Talbi
\thanks{University of Lille and INRIA, France, e-mail: el-ghazali.talbi@univ-lille.fr.}
}

\maketitle

\begin{abstract}
% XMVOPs --> Fixed MVOPs (CnOP, DOP)
Many real world optimization problems are formulated as mixed-variable optimization problems (MVOPs) which involve both continuous and discrete variables. MVOPs including dimensional variables are characterized by a variable-size search space. Depending on the values of dimensional variables, the number and type of the variables of the problem can vary dynamically. MVOPs and variable-size MVOPs (VMVOPs) are difficult to solve and raise a number of scientific challenges in the design of metaheuristics.

\medskip

Standard metaheuristics have been first designed to address continuous or discrete optimization problems, and are not able to tackle (V)MVOPs in an efficient way. The development of metaheuristics for solving such problems has attracted the attention of many researchers and is increasingly popular. However, to our knowledge there is no well established taxonomy and comprehensive survey for handling this important family of optimization problems. 

\medskip

This paper presents a unified taxonomy for metaheuristic solutions for solving (V)MVOPs in an attempt to provide a common terminology and classification mechanisms. It provides a general mathematical formulation and concepts of (V)MVOPs, and identifies the various solving methodologies than can be applied in metaheuristics. The advantages, the weaknesses and the limitations of the presented methodologies are discussed. The proposed taxonomy also allows to identify some open research issues which needs further in-depth investigations.
\end{abstract}

%\begin{keywords}
% Metaheuristics, Mixed-variable optimization problem, Variable-size mixed-variable optimization problem, Mixed optimization, Mixed variable programming, Variable-space design, Decomposition-based optimization.
%\end{keywords}

%************************
\section {Introduction}
%************************

% CnOP, DOP -->
Decision variables can be categorized into continuous and discrete variables. Continuous variables refer to real numbers defined within a given interval. Discrete variables can be divided into two categories: ordinal (i.e. integer, quantitative) and categorical (i.e. qualitative, nominal). According to the involved type of variables, optimization problems can be classified into continuous optimization problems (CnOPs) and discrete optimization problems (DOPs).

\medskip

However, many modern real world optimization problems (e.g. engineering design, process synthesis, logistics and transportation, bio-medical, machine learning \cite{talbi2021automated}\cite{meng2021comparative}) are formulated as mixed-variable optimization problems (MVOPs)\footnote{Also known in the literature as mixed variable programming (MVP).}. MVOPs contain both continuous and discrete variables. The most popular families of MVOPs are:
\begin{itemize}
	% MIP - MILP - MINLP
	\item {\bf Mixed-Integer Programming (MIP):} it represents a popular family of problems in mathematical programming with various domains of application such as logistics, transportation, manufacturing, energy, finance \cite{lodi2010mixed}. According to the linearity of objectives and constraints, one can classify this family into {\it Mixed-Integer Linear Programming (MILP)} or {\it Mixed-Integer Non-Linear Programming (MINLP)} problems. MIP problems are generally non convex and NP-hard \cite{cooper1981survey}. The challenging difficulty of MINLP problems is their non-linearity and non-differentiability due to the mixed decision space.
	
	% Black-box
	\item {\bf Black-box MVOPs:} they represent MVOPs problems in which the objective function and/or constraints cannot be formulated analytically. Moreover, they are generally characterized by expensive objective functions (e.g. intensive simulations). For instance, many engineering design (e.g. aerospace \cite{pelamatti2021mixed}, automotive, nanotechnology, mechanics) and machine learning problems (e.g. neural architecture search) are black-box MVOPs \cite{martins2021engineering}. 
\end{itemize}

\medskip

% Variable VMVOP - Dimensional variables
MVOPs including dimensional variables are characterized by a variable-size decision space. This family of variable-size MVOPs (VMVOPs) is referred in the literature as mixed-variable optimization problems \cite{lucidi2005algorithm} or variable-size design space problem \cite{nyew2015variable}\cite{abdelkhalik2021algorithms}. Depending on the values of dimensional variables, the number and type of variables of the problem can vary dynamically. Dimensional variables can also influence the number and type of constraints a given solution is subject to. For instance, in engineering design problems, dimensional variables can represent the choice between several possible sub-system components. Each component is usually characterized by a partially different set of design variables, and therefore, depending on the considered choice, different continuous and discrete design variables must be optimized. In the automated design of deep neural networks, the architectures are composed of hundreds of mixed variables: continuous (i.e. learning rate), discrete (e.g. number of layers) and dimensional (e.g. type of layer: convolution, pooling, fully-connected layer) \cite{talbi2021automated}.

\medskip

Designing efficient metaheuristics to solve complex (V)MVOPs\footnote{Family of optimization problems including VMVOPs and MVOPs} raises several challenges. Indeed, this family of problems are difficult to solve by standard metaheuristics according to the following distinguishing characteristics:
\begin{itemize}
	\item {\bf Mixed continuous and discrete variables}: most metaheuristics have been initially designed for CnOPs or DOPs. For example, standard differential evolution (DE), evolution strategies (ES), cuckoo search (CS), firefly algorithm (FA) have been initially developed for fixed-size CnOPs, while particle swarm optimization (PSO), ant colony optimization (ACO), simulated annealing (SA), and tabu search (TS) have been developed first for handling fixed-size DOPs. Hence, their application to (V)MVOPs need new methodologies for metaheuristics design.
	
	\item {\bf Categorical variables}: they represent elements (e.g. colors, material type) which are constrained to take values from a finite set of non-numerical and unordered values. They can be encoded as a set of discrete numbers. However, their relaxation results in non-computable objective functions and constraints, and they cannot be assigned intermediate values. 
	
	\item {\bf Dimensional variables}: they can be represented by a vector of integer values and induce VMVOPs. Indeed, each value of a dimensional variable has an impact on the number and type of the involved variables, the number of constraints and the structure of the objective function. Moreover, the discreteness of dimensional variables cannot be relaxed using continuous variables.
	
	\item {\bf Constraint handling}: in general, the integer restriction in (V)MVOPs divides the feasible region into discontinuous feasible sub-regions of differents sizes. Thus, constraint handling in (V)MVOPs represents a difficult issue to handle during the search.
\end{itemize}

\medskip

% Related work
The development of metaheuristics for solving (V)MVOPs has attracted the attention of many researchers and is increasingly popular. However, to our knowledge there is no well established taxonomy and comprehensive survey for handling this important family of optimization problems. Many papers concerned by MVOPs focus on a specific methodology for a given metaheuristic such as DE \cite{lampinen1999mixed}\cite{ho2015improved}\cite{zhao2020ensemble}, ES \cite{emmerich2000fixed}\cite{back1991survey}\cite{miyagi2018well}, PSO \cite{dos2010gaussian}\cite{kim2020constrained}, ACO \cite{socha2004aco}\cite{rivas2017coordination}\cite{liao2014fixed}, estimation distribution algorithms (EDA)  \cite{wang2019estimation}\cite{zhou2015estimation}, genetic algorithms (GA) \cite{jalota2018genetic}\cite{liu2018efficient}\cite{maiti2006application}, TS \cite{exler2008tabu}\cite{MashinchiOP11}, SA \cite{zhang1993mixed}\cite{koken2018simulated}\cite{mohan1999controlled}, artificial bee colony (ABC) \cite{akay2021survey} and so on. Some papers deal with specific families of MVOPs such as those where continuous variables are linearly constrained \cite{lucidi2005algorithm}. Moreover, very few papers deals with VMVOPs. Developing efficient metaheuristics for VMVOPs is in its infancy. 

\medskip

% Contributions of this paper
This paper presents a unified taxonomy for general metaheuristic methodologies for solving (V)MVOPs in an attempt to provide a common terminology and classification mechanisms. It provides a general mathematical formulation and concepts of (V)MVOPs, and identifies the various solving methodologies for metaheuristics than can be applied. The advantages, the weaknesses and the limitations of the presented methodologies are discussed. The proposed taxonomy also allows to identify some open research issues which needs further in-depth investigations.

\medskip

% Plan
The paper is structured as follows. In section \ref{formulation}, general formulations and concepts of (V)MVOPs are provided. In section \ref{metaheuristics}, the main search components of metaheuristics (e.g. initialization, encodings of solutions, neighborhoods, variation operators, model construction) that have to be adapted for solving (V)MVOPs are highlighted in a unified and most general way. The following sections \ref{mvop} and \ref{mvmop} present the taxonomy, classify, detail and analyze the various metaheuristic methodologies for solving respectively MVOPs and VMVOPs (e.g. global versus decomposition-based, encoding adaptation versus variation operators adaptation, relaxation versus discretization, sequential versus nested versus co-evolutionary). Finally, the last section \ref{conclusion} presents the main conclusions and opens some important research perspectives.

%*****************************************
\section {Mixed-variable optimization problems}
%*****************************************
\label{formulation}

In this section, a general mathematical formulation of MVOPs is provided, and is followed by the formulation of VMVOPs.

%***************************************************
\subsection {Fixed-size mixed-variable optimization problems}
%***************************************************

% Definition and characteristics
Fixed-size MVOPs are concerned by three different types of variables: continuous, discrete ordinal, and discrete nominal \cite{audet2001pattern}. Continuous and discrete ordinal variables are related to measurable values and by consequence, a relation of order between the possible values can be defined. Discrete nominal variables (i.e. categorical) are non-relaxable variables defined within a finite set of choices, in which no relation of order can be defined between the possible values.

\medskip

% Mathematical formulation
Let us introduce a fixed-size vector of mixed decision variables $(x,y)$, where $x \in \mathbb{R}^{n_x}$ is a vector of $n_x$ continuous variables, $y \in \mathbb{Z}^{n_y}$ is a vector of $n_y$ discrete variables. The discrete vector $y=(y_o,y_c)$ contains both ordinal and categorical variables. A MVOP can be formulated as follows (Eq.\ref{vmop}): 

\begin{equation} 
\label{vmop}
\begin{aligned}
\text{min } f(x,y)\text{, } & f: \mathcal{F}_x \times \mathcal{F}_y \longrightarrow \mathbb{R}  \\   
\text{subject to }  & x \in \mathcal{F}_x \subseteq \mathbb{R}^{n_x} \\
& y \in \mathcal{F}_y \subseteq \mathbb{Z}^{n_y} \\ 
% & z \in \mathcal{F}_z = \prod_{i=1}^{n_z}{\{ z_{i,1},..., z_{i,|C_i|} \}} \\ 
\end{aligned}
\end{equation} 

%\end{equation}
where $f(.)$ is the objective function, $\mathcal{F}_x$ (resp. $\mathcal{F}_y$) are the feasible region for continuous (resp. discrete) variables. The constraints are generally represented by bounding constraints $x=\prod_{i=1}^{n_x}{[x^{min}_i, x^{max}_i]}$, $y=\prod_{i=1}^{n_y}{[{y}^{min}_i,{y}^{max}_i]}$, and inequality constraints $g(x,y) \leq 0, g_i : \mathcal{F}_{x_i} \times \mathcal{F}_{y_i} \longrightarrow \mathcal{F}_{g_i} \subseteq \mathbb{R}  \text{ for } i=1,...,n_g$, where $n_g$ is the number of constraints.
% $|C_i|$ is the number of categorical values for the variable $z_i$

\medskip

% Mathematical properties
The feasible decision space of a MVOP can be decomposed into separated subsets according to the discrete vector $y$. For any fixed discrete vector $y^*=(y^*_1,..., y^*_{n_y})$, the set $A_{y^*}= \{ (x,y): x \in \mathcal{F}_x, y=y^* \} $ defines an $n_x$-dimensional affine sub-space. These affine sub-spaces for different values of $y^*$ are mutually disjoint. The distance $\delta$ between $A_{y_1}$ and $A_{y_2}$, $\delta = min \{ d((x,y_1), (x^{'} , y_2) ), \forall (x,y_1) \in A_{y_1}, \forall (x^{'} ,y_2) \in A_{y_2} \}$ is greater or equal than 1 \cite{cheung1997coupling}. Hence, the minimization of $f$ on $\cup_{y \in \mathcal{F}_y} A_{y}$ can be reduced as: $$ Min_{y \in \mathcal{F}_y} [ \, Min_{(x,y) \in A_y} f(x,y) ] \, $$ 
An exhaustive solving of the sub-problems for all $y \in \mathcal{F}_y$ is impractical, since the size of the set $\mathcal{F}_y$ is generally large.

%*****************************************************************
\subsection {Variable-size mixed-variable optimization problems}
%*****************************************************************

VMVOPs include dimensional variables which are non-relaxable variables defined within a finite set of choices. The size of the decision space and the definition of the constraint functions vary dynamically during the optimization process as a function of the values of dimensional variables \cite{lucidi2005algorithm}\cite{pelamatti2017deal}. Let us introduce $z$ as a vector of $n_z$ dimensional variables, a general VMVOP can be formulated as follows (Eq.\ref{vmvop}):
\begin{equation} 
%\begin{align*}
\label{vmvop}
\begin{aligned}
\text{min } f(x,y,z)\text{, } & f: \mathcal{F}_x \times \mathcal{F}_y \times \mathcal{F}_z \longrightarrow \mathbb{R} \\  
\text{subject to } & x \in \mathcal{F}_x(z) \subseteq  \mathbb{R}^{n_x(z)} \\ 
& y \in \mathcal{F}_y(z) \subseteq \mathbb{Z}^{{n_y(z)}} \\ 
& z \in \mathcal{F}_z \subseteq \mathbb{Z}^{{n_z}} \\
\end{aligned}
%\end{align*}
\end{equation}
where $x$ and $y$ are variable-size vectors of size respectively equal to $n_x(z)$ and $n_{y}(z)$. The number of constraints $n_g(z)$ the problem is subject to is also variable: $g(x,y,z) \leq 0, g_i : \mathcal{F}_{x_i}(z) \times \mathcal{F}_{y_i}(z) \times \mathcal{F}_z \longrightarrow \mathcal{F}_{g_i} \subseteq \mathbb{R}, for \hspace{0,2cm} i=1,...,n_g(z)$. 

\medskip

% Neighborhoods: Continuous neighorhood (ball) and discrete neighborhood
The concept of global optimality is straightforward to generalize for VMVOPs.
\begin{definition}
	\label{local-optimum}
	A solution $(\widehat{x},\widehat{y},\widehat{z})$ is a global optimum if $f(\widehat{x},\widehat{y},\widehat{z}) \leq f(x,y,z), \forall z \in \mathcal{F}_z, \forall y \in \mathcal{F}_y(z), \forall x \in \mathcal{F}_x(z)$.
\end{definition}

% Local optimum solution
Local optimality for VMVOPs is less obvious to define. One has to take into account simultaneously both discrete and continuous variables \cite{audet2001pattern}. Modifying the discrete variables make sense only if the continuous variables change as well. Indeed, variations of the discrete variables will involve changes in the continuous neighborhoods. A neighborhood $\mathcal{N}_c(x)$ for continuous variables $x$ is represented by a ball $\mathcal{B}(\epsilon,x)$ of radius $\epsilon$ around $x$. Local optimality can be defined as: $x^*$ is a local minimum if $ f(x^*) \leq f(x)$ for all $x \in \mathcal{B}(\epsilon,x^*)$. The same definition holds for discrete variables using combinatorial finite neighborhoods which will depend on the nature of the variables of the VMVOPs to solve (e.g. binary, bounded ordinal, categorical, permutation). The discrete feasible finite neighborhood $\mathcal{N}_d(x,y,z)$ take into account the fact that changes of the discrete variables can also have an impact on the values of the continuous variables. Local optimality in VMVOPs is then related to the balls centered at the solutions belonging to the discrete neighborhood. A solution $(x^*,y^*,z^*)$ is a local optimum if there are no better solutions in the continuous neighborhood of all the solutions belonging to its discrete neighborhood. 

\begin{definition}
	Given a discrete neighborhood $\mathcal{N}_d(x,y,z)$, a solution $(x^*,y^*,z^*)$ is a local optimum if for all $(\overline{x},\overline{y},\overline{z}) \in \mathcal{N}_d(x^*,y^*,z^*)$, $f(x^*,y^*,z^*) \leq f(x,\overline{y},\overline{z}), \forall x \in \mathcal{B}(\epsilon,\overline{x}) \cap \mathcal{F}_x(\overline{z})$, $\epsilon > 0$.
\end{definition}

% Stationary solution
% Finding a local optimum solution for VMVOPs may be prohibitive. It is more reasonable to find a {\it stationary solution} of the continuous problem generated by fixing the the values of the discrete variables. 
%\begin{definition}
%Given a feasible finite discrete neighborhood $\mathcal{N}_d(y^*,z^*)$, A solution $(x^*,y^*,z^*)$ is a stationary solution if (i) it is a stationary solution of the following continuous optimization problem: $\text{min } f(x,y^*,z^*), x \in \mathcal{F}_x(z^*)$, (ii), (iii).
%\end{definition}

%**************************************************************
\section{Main metaheuristic components}
%**************************************************************
\label{metaheuristics}

% Metaheuristics
Metaheuristics represent a class of general-purpose approximate algorithms that can be applied to difficult optimization problems \cite{Talbi2009}. One can classifiy metaheuristics into {\it local-search} based and {\it population-based metaheuristics}. Local search-based metaheuristics improve a single solution. They could be seen as search trajectories through neighborhoods. They start from a single solution. They iteratively perform the generation and replacement procedures from the current solution $s$. In the generation phase, a set of candidate solutions are generated from the neighborhood $\mathcal{N}(s)$ of the current solution. In the replacement phase\footnote{Also named transition rule, pivoting rule and selection strategy.}, a selection is performed from the candidate solution set to replace the current solution (i.e. a solution $s^{'} \in \mathcal{N}(s)$) is selected to be the new solution. Popular examples of such metaheuristics are local search (LS) (i.e. hill-climbing, gradient), simulated annealing and tabu search.

\medskip

Population-based metaheuristics could be viewed as an iterative improvement of a population of solutions. They start from an initial population of solutions. Then, they iteratively apply the generation of a new population and the replacement of the current population. Using various bio-inspired metaphors, they may be classified into two main categories \cite{Talbi2009}\cite{del2019bio}: 
\begin{itemize}
	% Evolutionary algorithms
	\item {{\bf Evolutionary algorithms (EAs):} they are guided by the selection and reproduction concepts of survival of the fittest. The solutions composing a population of solutions (i.e. individuals) are selected and reproduced using variation operators (e.g. mutation, crossover\footnote{Also called recombination and merge.}). New offsprings (i.e. solutions) are constructed from the different features of solutions belonging to the current population. The most popular EAs are genetic algorithms, differential evolution \cite{storn1997DE}, evolution strategy, and estimation distribution algorithms.
	}
	% Swarm intelligence
	\item {{\bf Swarm intelligence (SI):} they are inspired by the emergence of collective intelligence from populations of agents with simple behavioral patterns for cooperation \cite{eberhart2001swarm}. Particle swarm optimization, ant colony optimization, firefly algorithm, and cuckoo search represent examples of this class of metaheuristics.
	}
\end{itemize}
% Regular- model-based
Population-based metaheuristics can also be categorized into {\it regular population} and {\it model-based population} algorithms \cite{stork2020new}. They are distinct in the way the offsprings are generated during the search. In regular population algorithms, the solutions are generated using population-based variation operators (e.g. mutation and crossover for GA, DE, ES, and particle update for PSO), whereas in model-based algorithms, the solutions are generated by using and adapting a model which stores information during the search (e.g. pheromone in ACO, statistical model in EDA). 

\medskip

% Important for ALL meta: encodings and intialization
The common search components for all metaheuristics which have to be adapted to solve (V)MVOPs are the {\it encoding} of solutions and their {\it initialization}. Indeed, the representation of mixed and variable-size solutions is a crucial issue and has an important consequence on the efficiency of metaheuristics. Indeed, it defines the decision space in which any metaheuristic will carry out the search process. In general, the initial solutions are randomly and uniformly generated according to the type of variables and the constraints associated to the (V)MVOP. Continuous (resp. ordinal) variables are generally defined by bounded intervals, while categorical and dimensional variables are defined by a finite domain of possible values.

\medskip

% Important for LS-based
% Common concepts for local search-based metaheuristics
In addition to encodings, another important search component to be adapted for local search-based metaheuristics is the definition of the neighborhood and its exploration. Finding a local optimal solution for (V)MVOPs (e.g. expensive black-box) according to the neighborhood definition given in Eq.\ref{local-optimum} can be prohibitive. Concerning population-based metaheuristics, the most important search components involved for solving (V)MVOPs are the variation operators for regular population-based  (e.g. GA, DE, PSO) and the model construction and update for model-based algorithms (e.g. ACO, EDA).

% In practice, it is more reasonable to determine a point (usually called stationary point) which satisfies suitable necessary optimality conditions \cite{lucidi2005algorithm}.

%*************************************************
\section{A taxonomy of metaheuristics for MVOPs}
%*************************************************
\label{mvop}

\begin{figure*}[!t]
	\centering
	\includegraphics[width=0.8\textwidth]{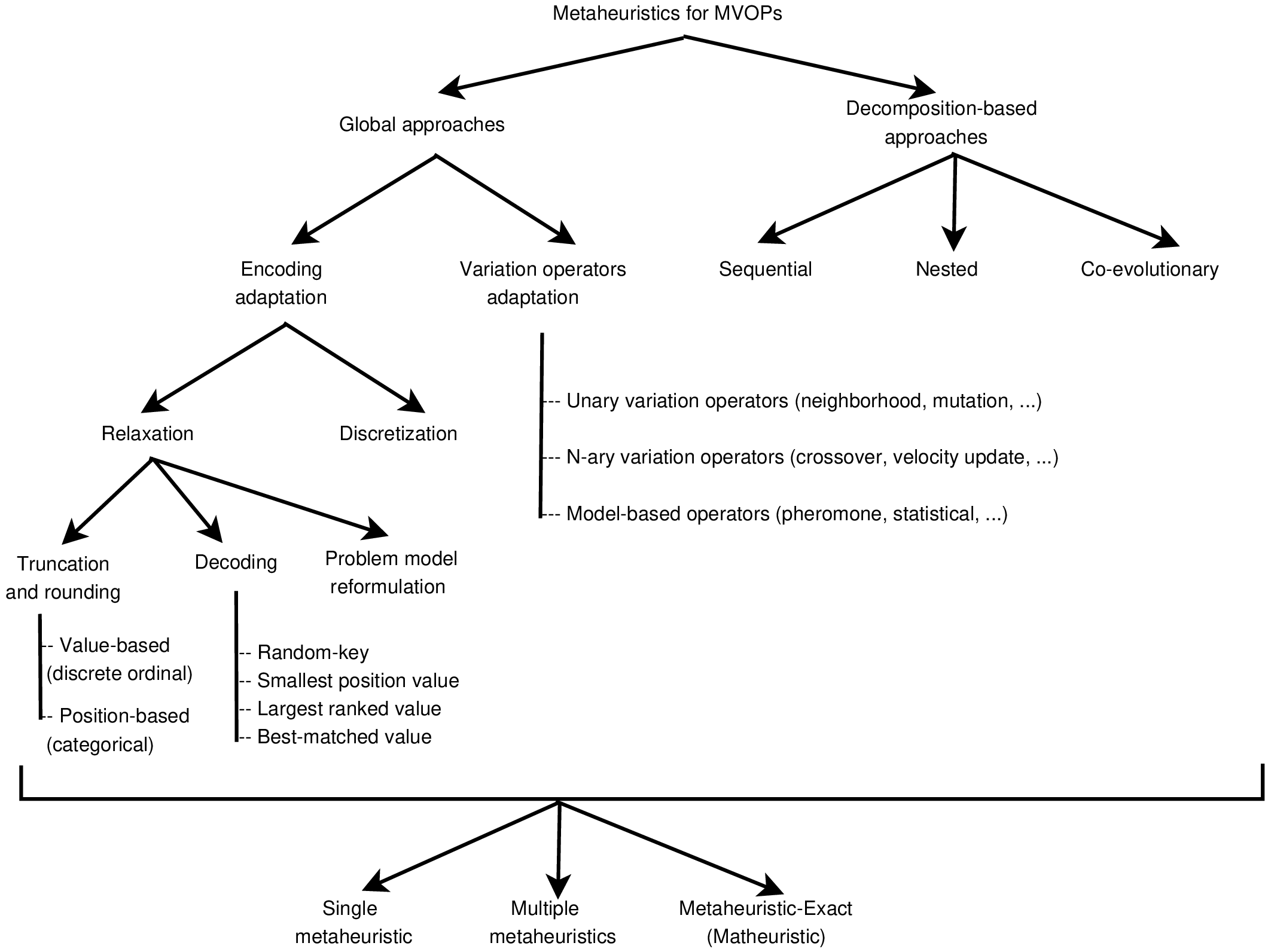}
	\caption{A taxonomy of metaheuristic methodologies for solving MVOPs.}
	\label{fig:taxonomy}
\end{figure*}

The proposed taxonomy is based on a unifying view of metaheuristics in solving MVOPs according to the main concerned search components. Using this taxonomy, a survey describes and discusses the various general metaheuristic methodologies to solve MVOPs and gives some relevant illustrative algorithms. Solving methodologies that require problem-specific knowledge of MVOPs are not considered in this paper \cite{nyew2015variable}. The first criteria of our taxonomy is related to the definition of the decision space in which metaheuristics will operate. First, the algorithms can be categorized in two types (Fig.\ref{fig:taxonomy}):
\begin{itemize}
	\item {\bf Global approaches:} the optimization process of metaheuristics is carried out in the whole mixed decision space. 
	\item {\bf Decomposition-based approaches:} some partitioning of the fixed-size mixed decision space is performed. According to a given decomposition, a set of subproblems are generated and solved. 
\end{itemize}

%******************************************************************
\subsection{Global approaches for MVOPs}
%******************************************************************

Global approaches consist in applying a global search on the whole decision space of mixed variables. The discrete and the continuous variables are treated simultaneously. This approach maintains the relationships between the decision variables, and involves a straightforward design and implementation. The goal here is the reuse of metaheuristics designed for CnOPs or DOPs to be easily deployed to solve MVOPs. However, the features of standard metaheuristics, and the consistency and compatibility of encodings and variation operators can be destroyed. The global approach contains two main categories: {\it encodings adaptation} and {\it variation operators adaptation}.

%******************************************************************
\subsubsection{Encodings adaptation}
%******************************************************************

In a MVOP, a solution is generally represented by a fixed-size vector of mixed variables $(x,y)$ where $x$ (resp. $y$) represents the continuous (resp. discrete variables). The most popular approach consists in mapping completely the encoding of a MVOP into a continuous or discrete encoding. The two main mapping methods are {\it relaxation} and {\it discretization}. 

\medskip

{\bf Relaxation:} this approach consists in relaxing the discrete variables as continuous ones. Then, a MVOP is transformed into a CnOP, and any continuous metaheuristic can be applied \cite{yildiz2020comparative}\cite{gupta2021comparison}. Variation operators in those metaheuristics (e.g. neighborhoods, mutation, crossover, particle update) will generate real numbers. The conversion of real values into integer values is necessary just to evaluate the objective function and constraints, and then continuous representations are keeped in the search process. 

\medskip

{\it Conversion of discrete ordinal variables:} the most popular and simple conversion function for discrete ordinal variables is the {\it truncation} function $y_i = INT(y_i)$, where $INT(.)$ is a function modifying a real value into an integer one by rounding to the nearest available value. Metaheuristics based on truncation have been widely deployed in the literature to solve MVOPs:
\begin{itemize}
	% LS
	\item{\bf Local search-based algorithms}: it concerns mainly the continuous neighborhood definition, and has been applied to SA \cite{zhang1993mixed}\cite{koken2018simulated}\cite{mohan1999controlled}, TS \cite{MashinchiOP11}, and spiral dynamics optimization algorithm \cite{kania2016solving}.
	% EA
	\item{\bf Evolutionary algorithms}: truncation concerns the mutation and some crossover operations in DE \cite{lampinen1999mixed}\cite{ho2015improved}\cite{zhao2020ensemble}\cite{angira2006optimization} \cite{ali2020novel}\cite{zheng2019differential}\cite{mohamed2017efficient}\cite{ponsich2011differential}\cite{varadarajan2008differential}\cite{mohamed2019solving}\cite{lin2004mixed}\cite{peng2021multi}, and GA (e.g. arithmetic crossover, non-uniform mutation) \cite{rao2005hybrid}\cite{deep2009real}\cite{wu1995genetic}\cite{jalota2018genetic}\cite{liu2018efficient} \cite{maiti2006application}\cite{yokota1996genetic}~\cite{yan2004solving}. Truncation is unnecessary for some crossover operators (e.g. uniform crossover) \cite{jun2013improved}\cite{lin2001co}.  
	% SI
	\item{\bf Swarm Intelligence}: truncation is used in the particle movement operation of PSO \cite{guo2004swarm}\cite{nahvi2011particle}\cite{chowdhury2013mixed}\cite{venter2004multidisciplinary}\cite{parsopoulos2002recent}\cite{dos2010gaussian}\cite{kim2020constrained}, CS \cite{kanagaraj2014effective}\cite{gandomi2013cuckoo}, harmony search (HS) \cite{mahdavi2007improved}, Grey-Wolf \cite{gupta2019efficient}, teaching-
	learning-based optimization (TLBO) \cite{rao2011teaching}) \cite{talatahari2020discrete}, ABC \cite{das2017transmission}, and FA \cite{gandomi2011mixed}. For ACO, various statistical models are applied to deal with different type of variables \cite{SchluterEB09}\cite{LiaoKH12}\cite{liao2014fixed}.
	% Hybrid meta
	\item {\bf Hybrid metaheuristics}: truncation can be extended to any hybrid continuous metaheuristic. Some combinations have been investigated, such as combining DE with ABC \cite{miao2016modified}, DE with HS \cite{liao2010two}, ACO with chaotic optimization \cite{dos2008use}, FA with chaotic optimization \cite{dos2011chaotic}, DE with TS \cite{srinivas2007differential}, DE with direct search \cite{yi2013three}, TLBO with HS \cite{talatahari2020discrete}, and GA with PSO \cite{garg2016hybrid}. 
\end{itemize}

\medskip

% Analysis of truncation: discontinous feasible regions - Flat landscape
Another popular conversion method is {\it rounding} (Fig.\ref{fig:taxonomy}). A better discrete solution is not always selected by rounding \cite{rao2019engineering}. Another conversion approach consists in checking the upper and lower discrete values of the continuous variables and selecting the best one \cite{nahvi2011particle}. This approach obtains better results (i.e. accuracy, convergence) especially when the range of discrete variables is important. However, it involves additional evaluations of the objective function, and then higher computational cost \cite{kitayama2006method}. Other more complex value-based relaxations can be used \cite{lotfipour2016discrete} such as the nearest vertex approach (NVA) \cite{chowdhury2013mixed} and the shortest normal approach (SNA) \cite{chowdhury2010developing}. 

\medskip

% Constraints, flat landscape
Relaxation has two main advantages. On one hand, it does not oblige the design of new effective  variation operators for discrete variables. On the other hand, the same variation operators are applied to the whole mixed vector. However, discrete variables generated by relaxation strategies do not always satisfy the constraints \cite{fu1991mixed}. Relaxation can also have an influence on entering the feasible regions of a MVOP which are characterized by discontinuous feasible regions. The search process tends to enter the largest feasible regions. Once entering a non-optimal large-size feasible region, it is difficult to escape towards small-size feasible regions. In some problems, the global optimal solutions may be localized in small-size feasible regions, which are neglected. 

\medskip

The encoding adaptation based on relaxation will lead to a biaised search procedure and potentially generates a harder optimization problem. The extended continuous search space contains a large number of infeasible solutions. Hence, the search process can generate solutions which are far from the feasible region, and needs a higher computational cost. Moreover, the transformed MVOPs are characterized by a huge number of local optima, which makes them more complex to solve by continuous metaheuristics \cite{murray2010algorithm}. Another drawback of relaxation is that it introduces flat areas in the fitness landscape of MVOPs, for which metaheuristic search is less efficient.

\medskip

\medskip

% Categorical variable and position-based
{\it Conversion of categorical variables:} relaxation strategies have also been applied to treat categorical variables. However, it does not work effectively for categorical variables for which there is no ordering relation between their values, and are not numerically related. There is no sense on carrying out arithmetic operations on categorical values. Hence, the relaxation of categorical variables using value-based conversion is less natural and may decrease the performance of metaheuristics as the number of categorical variables increases \cite{liao2014fixed}. For some MVOPs, the categorical values are associated to a set of continuous values which can be used during the search. {\it Position-based encoding} is more adapted to encode categorical variables \cite{abhishek2010modeling}. Let $Z=\{ z_1,...,z_i, z_n\}$ be a set values for a given categorical variable. One can use the position (i.e. index) of the discrete variable instead of the discrete value \cite{he2004improved}. Then, a continuous variable associated to a categorical variable can be divided into equal intervals which are associated to the index values of the categorical variable. Uniform-spacing integer positions from $1$ to $n$ are used and replaced by continuous position variables. Then, the truncation and rounding strategies can be applied for such position-based encodings. For instance, This approach has been investigated in DE \cite{lampinen1999mixed} and threshold accepting (TA) algorithms \cite{schmidt2005combined}. This strategy is based on the assumption that neighboring categorical values contribute similarly to the objective function, which is not always the case. In many categorical variables, the values define qualitative distinctions. 

\medskip

This approach has also been adapted for binary variables \cite{angira2006optimization}\cite{crawford2017putting}. For instance, one can use sigmoid-based conversion by following a logistic probability distribution: $ z = 1 \text{ if } r \leq P(x), 0 \text{ otherwise }$ where $P(x)=\frac{1}{1+e^{-x}}$, $r$ is a random term following a uniform distribution in the interval $[0,1]$, $z$ is binary variable, and $x$ is the continuous variable to transform. This approach has been integrated in DE \cite{yu2016stock}, PSO \cite{kennedy1997discrete} and FA \cite{palit2011cryptanalytic}. Other functions such as trigonometric function can also be applied \cite{pampara2006binary}.

\medskip

% Decoding - Random-key
{\it Conversion of order-based encodings:} truncation and rounding cannot be applied to some discrete encodings such as order-based encodings (e.g. permutation). An increasingly popular relaxation method is based on {\it decoding}. The most popular decoding approach is the {\it random-key representation} \cite{LiuWJ07}. It transforms a position in a continuous space into a position in a discrete space. It has been used first for permutation-based DOPs (e.g. scheduling \cite{LiuWJ07}\cite{ali2016differential}, routing \cite{ouaarab2015random}, assignment \cite{hafiz2016particle}). Random-key encodings have been used in most of the metaheuristics such as PSO \cite{LiuWJ07}\cite{ouaarab2015random} and DE \cite{ali2016differential}. Other decoding methods have been used in the literature such as the smallest position value (SPV) rule into PSO \cite{tasgetiren2007particle}, HS \cite{ulker2016adaptation}, FA \cite{yousif2014discrete}, CS \cite{kumar2011design}, the largest ranked value (LRV) rule \cite{li2013hybrid}, and the best-matched value (BMV) rule \cite{ali2020novel}.

\medskip

% Problem model reformulation
Handling discrete variables in relaxation approaches can also be handled through the {\it problem model reformulation}. It is generally carried out by penalty constraints through which the correct values of discrete variables cannot be reached. A classical relaxation method for categorical variables encodes them by continuous ones and includes some constraints in the model. This is a popular reformulation methodology in mathematical programming \cite{abhishek2010modeling}\cite{arora1994methods}\cite{davydov1972application}\cite{shin1990penalty}. For instance, the addition of the following non-convex nonlinear function $z(1-z)=0, 0 \leq x \leq 1$ to model a binary variable will force $z$ to take either $0$ or $1$ \cite{Li92}. In \cite{kitayama2006penalty}\cite{kitayama2006method}, using PSO, the authors incorporate a dynamic penalty approach into the objective function to handle discrete variables as continuous variables.  

% Additional objectives

\medskip

% Discretization
{\bf Discretization approaches:} this mapping function consists in discretizing the search range of continuous variables so that they can be optimized as discrete variables. It facilitates the use of standard metaheuristics and variation operators originally developed for DOPs:
\begin{itemize}
	% GA
	\item{\bf Genetic algorithms:} it has been applied in simple GAs which encode all continuous variables as binary strings \cite{GoldbergDC92}\cite{michalewicz1996genetic}~\cite{dimopoulos2007mixed}\cite{costa2001evolutionary}\cite{turkkan2003discrete}\cite{lin1992genetic}. Thus, the standard binary crossover and mutation variation operators can be applied on continuous variables. 
	% EDA
	\item{\bf Estimations of distribution algorithms:} it is possible to transform each continuous parameter into a discrete one and then use classical statistical models such as Bayesian networks. However, with the increasing precision of continuous variables, the complexity of the statistical models grows exponentially.
	% SA 
	\item {\bf Local search-based algorithms:} the discretization of the variables facilitates the definition of a discrete neighborhood. Then, any discrete local search-based metaheurictic such as SA \cite{stelmack1998genetic} can be applied to solve the MVOP as a fully discrete one.
\end{itemize}

% Analysis
This approach is limited in terms of accuracy. For some MVOPs, continuous variables require high accuracy, and the precision of continuous variables is restricted and suffers from limited solution accuracy. For instance, using the binary discretization, the performance of the metaheuristics may be decreased by the high dimensionality of the binary decision space especially when high accuracy is required.

\medskip

% Inconvenient general de ces methodes
Although the methodology based on encoding adaptation is easy and simple to design and implement, its efficiency can suffer from losing important knowledge by applying successive mappings of a continuous (resp. discrete) space into a discrete (resp. continuous) space.

%******************************************************************
\subsubsection{Variation operators adaptation}
%******************************************************************

This approach does not apply any transformation of the mixed-variable encoding scheme. Instead of modifying the encoding, new variation operators are introduced to handle the type(s) of decision variables that cannot be handled originally by a given metaheuristic \cite{Coelho10}\cite{wang2008ranking}. It is based on the adaptation of the variation operators of standard metaheuristics to handle both continuous and discrete variables separately (see Fig.\ref{fig:mutationoperator} and Fig.\ref{fig:variationoperator}). In this combined variation method, the continuous and the discrete variation operators are carried out simultaneously to generate new solutions. Different variation operators will handle different type of variables, and then there is no more restriction to deal with mixed-variable encodings. 

\medskip

Three main design questions arize in the adaptation of variations operators: 
\begin{itemize}
	\item The definition of the variation operators to handle the different types of variables.
	\item The selection of the variation operators to apply at each iteration of the algorithm.
	\item The selection of the variables on which to carry out the variation operator. 
\end{itemize}

\medskip

According to the various families of variation operators, the following methodologies have been investigated:
\begin{itemize}
	% Neighborhood, mutation
	\item{\bf Unary operators (e.g. neighborhhood, mutation):} 
	% Neighborhood
	the main issue in local search-based algorithms is the handling of the large mixed neighborhood and the exploration of the various neighborhoods. One can use a complete generation of the mixed neighborhood and applying a first improvement strategy such as proposed in \cite{de2019hybridizing} for variable neighborhood search (VNS). However, it is impracticable to consider all neighbors of all neighborhoods. Deterministic or random selection of neighborhoods may be applied. In \cite{gardi2011local}, a local search (LS) algorithm is proposed, in which combinatorial and continuous neighborhoods are treated separately using a random stategy. A random generation of neighbors is carried out for both neighborhoods. In \cite{CardosaSA}, at each iteration a random selection of neighborhood is performed. Then, a SA algorithm is applied for the discrete neighborhood and a non-linear simplex of Nelder and Mead \cite{olsson1975nelder} is applied for the continuous neighborhood. In \cite{exler2008tabu}, a TS algorithm is proposed, in which the mixed neighborhood is dynamically partitioned into equal-size regions, and a set of random neighbor solutions are generated.
	
	\medskip
	
	% Mutation
	For the mutation operator in EAs, various mutations for continuous, discrete ordinal and categorical variables can be defined and simultaneously applied in a probabilistic way \cite{rudolph1994evolutionary}\cite{van2016multicriteria}\cite{li2014discrete}\cite{li2013mixed} (Fig.\ref{fig:mutationoperator}). In \cite{lin2018hybrid}\cite{datta2013real}, a combination of continuous DE with discrete DE is proposed. The continuous variables are initialized and evolved through the original DE, whilst the discrete variables are initialized and evolved through a discrete DE. In \cite{li2008fixed}, the authors combines mutation operators of ES in the continuous domain \cite{beyer2002evolution}, integer domain \cite{rudolph1994evolutionary}, and binary domain \cite{back1996evolutionary}. In opposition to mutation operators in standard ES, the mutation operators for mixed-integer spaces works with a heterogeneous distribution, combining the $l2$ symmetric normal distribution for continuous variables, a $l1$ symmetric geometric distribution for discrete ordinal variables, and a uniform distribution for the categorical values. The same methodology has been applied to CMA-ES \cite{miyagi2018well}, EP \cite{CAO2000931} and GA \cite{kincaid2004bell}. 
	
\begin{figure}[!t]
			\centering
			\includegraphics[width=2.5in]{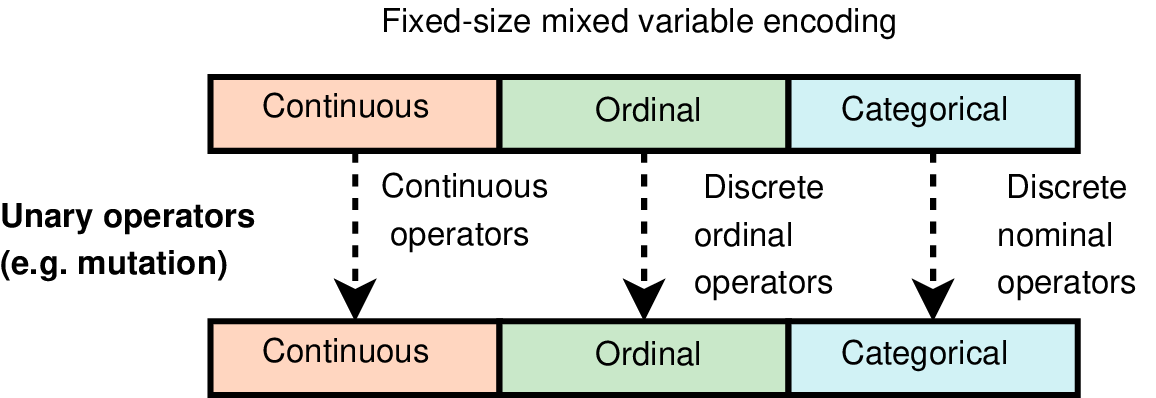}
			\caption{Adaptation of mutation operators in global approaches for handling MVOPs.}
			\label{fig:mutationoperator}
\end{figure}
	
	% EA
	% The probability of mutation is a parameter for each discrete variable
	
	%  Crossover, particle update
	\item{\bf N-ary operators (e.g. crossover, particle update):} the different standard N-ary variation operators for continuous and discrete variables are separately and independently applied to generate a part of the encoding. Then, the various parts of encodings are combined to construct a complete solution.
	% Crossover
	For the crossover operators in EAs, most of the standard operators can be used without any change (Fig.\ref{fig:variationoperator}). In \cite{li2014discrete}, the authors proposed an orthogonal crossover combined with the DE crossover. The same principle has been applied in ES \cite{emmerich2000fixed}, in which the standard dominant recombination (resp. intermediate) crossover operators is performed to discrete (resp. continuous) vectors \cite{back1991survey}.
	
	\begin{figure*}[!t]
	\centering
			\includegraphics[width=0.8\textwidth]{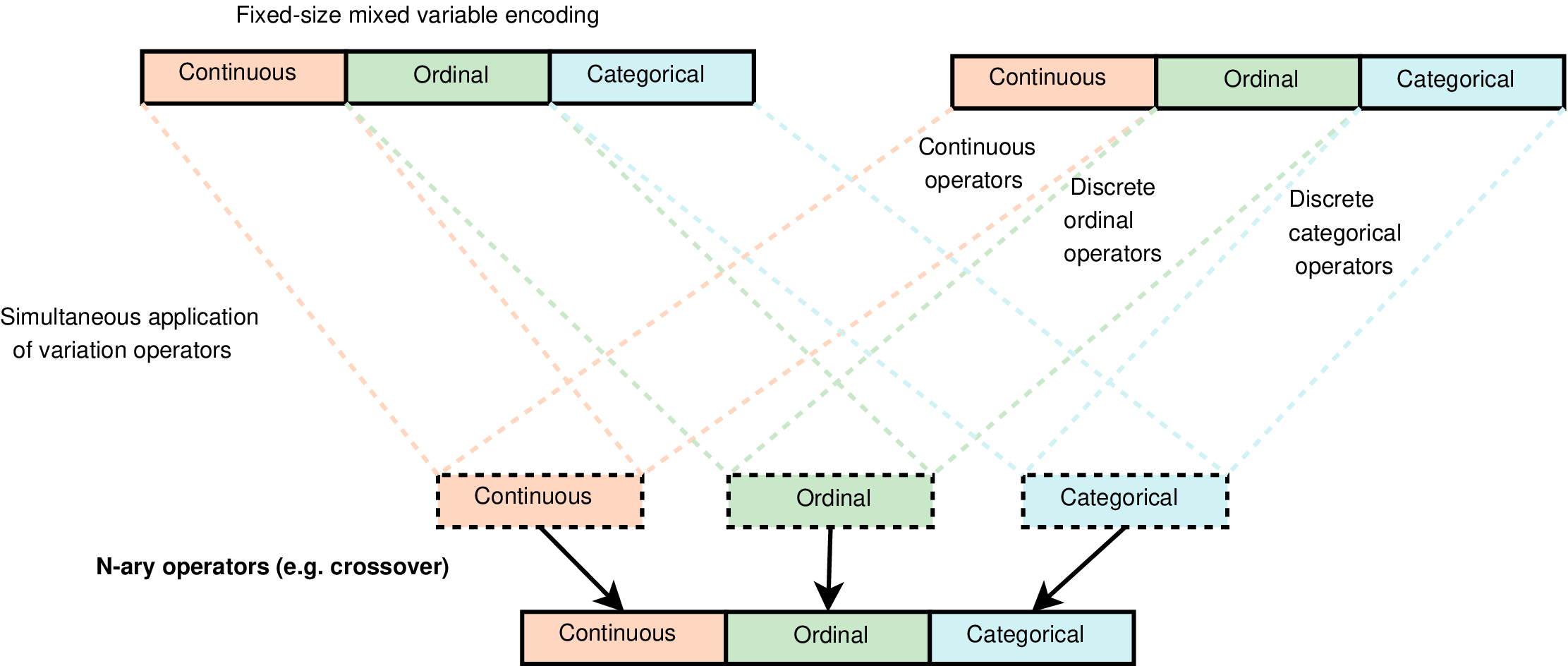}
			\caption{Adaptation of crossover operators in global EA approaches for handling MVOPs.}
			\label{fig:variationoperator}

	\end{figure*}
	
	\medskip
	
	% Particle update
	In the particle update operator of PSO, various operators, respectively for continuous and discrete variables, are simultaneously applied to the mixed vector \cite{gao2010comprehensive}\cite{wang2021particle}\cite{mokarram2018new}. Specific operators have been designed for discrete variables \cite{hinojosa2013modeling}\cite{sun2011modified}, binary  \cite{rezaee2020mixed}\cite{gao2010comprehensive}, and categorical variables \cite{liao2007fixed}\cite{dimopoulos2007mixed}. A random selection of particle update operators in PSO is performed (i.e. continuous or discrete) in \cite{yiqing2007improved}. Different priorities can be assigned for variables. In \cite{wang2008ranking}, a higher priority is associated to categorical variables. Among categorical variables, a random selection is performed. Subsequently, the discrete variable are prioritized as a function of both their performance and their feasibility with the constraints. 
	
	\medskip
	
	Other swarm intelligence metaheuristics have been investigated for MVOPs. In \cite{wang2017hybrid}, the authors adapt the HLO metaheuristic (Human Learning Optimization) in which real-coded parameters are optimized by the continuous linear learning operators of continuous HLO while the rest variables of problems are handled by the binary learning operators of HLO.
	
	\medskip
	
	%  Model-based search - construction and update
	{\bf Model-based operators (e.g. pheromone, statistical model):} in model-based metaheuristics such as ACO and EDA, the concerned operation is the construction and the update of the model.
	% ACO
	In ACO, one has to extend the initialization and update of the pheromone model for CnOPs. In DOPs, the ants construt a solution by making a probabilistic decision according to some discrete probability distribution (i.e. pheromone). In case of CnOPs, the domain changes from discrete to continuous. Instead of a discrete probability distribution, the Probability Density Function (PDF) is used \cite{socha2004aco}. The ants generate a random number according to a certain PDF $P_i(x_i)$. The construction for continuous variables is followed by discrete variables \cite{rivas2017coordination}. In \cite{liao2014fixed}, the continuous and ordinal variables are treated in a similar way as in the original ACO, while the categorical variables values are fixed through a random sampling which depends on the number and the performance of the solutions of the archive that use the given value.
	% EDA
	Concerning the EDA metaheuristic, the statistical models have been adapted to deal with continuous variables (e.g. Gaussian network \cite{larranaga1999optimization}, Gaussian kernel \cite{bosman2000mixed}, decision trees \cite{ocenasek2002estimation}). In \cite{wang2019estimation}, EDA has been extended by effective histogram models to handle mixed variables. In \cite{zhou2015estimation}, the authors propose a variable-width histogram to deal with continuous variables. The continuous space is decomposed into discrete bins, in which continuous variable are sampled. 
	% using a probabilistic model and generate new solutions by sampling from the model
\end{itemize} 

\medskip

% Hybrids - {\bf Heterogeneous variation operators:} Attention avec l'hybride sequentiel
The variation operators are not necessarily inherited from the same metaheuristic. Using heterogenous variation operators from different metaheuristics has been investigated in the literature. In \cite{sahoo2014efficient}, the authors propose an efficient hybrid approach based on GA and PSO combining integer and continuous variation operations of GA and PSO (mutation, crossover, particle update). In \cite{gao2016difference}, DE and GA variation operators are handling respectively the continuous and discrete parts. In \cite{hedar2011filter}, a GA is sequentially combined with pattern search. A set of elite solutions found by the GA are improved using pattern search.

\medskip

% Analysis - advantages & limitations
There are two main advantages using the variation operators adaptation methodology \cite{shi2017adaptive}\cite{wang2019estimation}. First, no transfomation of the mixed encoding is necessary to convert the variables, which generates higher computational cost. Second, different variation operators designed for various types of variables can be combined. Those variation operators are inherited from the same metaheuristic or not. Hence, this methodology does not need any major design change of metaheuristics. Many well-studied variation operators for CnOPs and DOP can be simply combined to solve MVOPs. Different probabilities can be assigned to the various variation operators \cite{li2021sizing}. Although this methodology has no limitations on the mixed type of variables, there is no guarantee that the variation operators for different types of variables are compatible and consistent. Therefore, this class of algorithms can be inefficient in guiding the search in mixed spaces.

%******************************************************************
\subsection{Decomposition-based approaches for MVOPs}
%******************************************************************

Decomposition approaches consist in partitioning the mixed decision space according to the type of variables. The discrete and the continuous variables are not treated simultaneously. This approach alternates iterations for which different types of variables are optimized. Decomposition strategies which use problem-specific knowledge of the MVOP are not considered in this section \cite{venter2004multidisciplinary}. There exists three main general decomposition-based approaches: {\it sequential}, {\it nested}, and {\it co-evolutionary}.

%******************************************************************
\subsubsection{Sequential approaches for MVOPs}
\label{sequential}
%******************************************************************

\begin{figure*}[!t]
\centering
\includegraphics[width=0.8\textwidth]{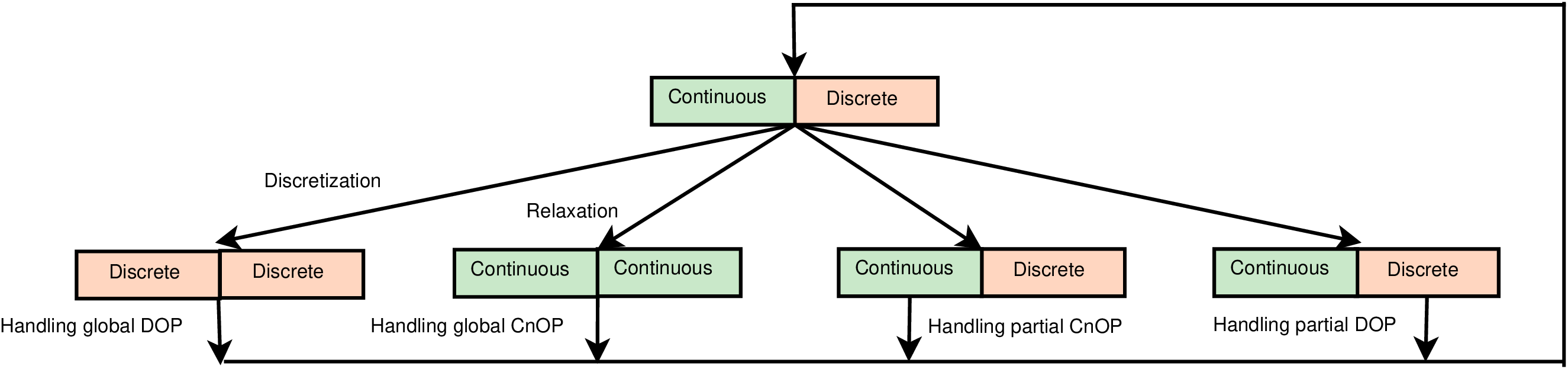}
\caption{Building blocks for decomposition-based sequential approaches in handling MVOPs.}
\label{fig:sequential}
\end{figure*}

In the sequential approach, the generated subproblems are solved in a sequential way. The subproblems are generated according to a given decomposition of the mixed decision space (Fig.\ref{fig:sequential}). Two design questions arise in this methodology:
\begin{itemize}
	\item{\bf Sequence of optimization:} this question concerns the order in which the different types of decisions variables are optimized.
	\item{\bf Complete or partial optimization:} in each sequential phase, the optimization process may concern a given partition of the decision space or a relaxation/discretization of the whole decision space.
\end{itemize}

% Continuous then discrete - Partial or global
According to those design questions, many sequential approaches may be designed to solve MVOPs. A popular approach consists in performing first an optimization of the continuous variables, followed by the optimization of the discrete ones:
\begin{equation}
\begin{aligned}
\text{min } f(x,y^*) \text{         } & \longrightarrow^{x^*} & \text{min } f(x^*,y)   \\
\text{w.r.t } x \in F_x   \text{        }    & \longleftarrow^{y^*}  & \text{w.r.t. } y \in F_y \\
\text{s.t. } g(x,y^*) \leq 0  &   & \text{s.t. } g(x^*,y) \leq 0
\end{aligned}
\label{sequential-vxop}
\end{equation}

\medskip

% Pattern search
In \cite{abramson2009mesh}\cite{abramson2004filter}\cite{abramson2003pattern}\cite{lucidi2005algorithm}, the common methodology, called {\it generalized pattern search}\footnote{Also called mesh-based local search.}, consists in alternating between a partial continuous and partial discrete search according to various neighborhoods. The size of the neighborhood is an important issue in this family of algorithms. In \cite{audet2006sequential}\cite{abramson2009mesh}, the neighborhood size varies during the search as a function of the quality of solutions obtained during the previous iterations . The continuous phase can also handle the whole decision variables by using relaxation. In \cite{capitanescu2010sensitivity}, the sequential approach handles a continuous optimization of the whole decision space by relaxing the subset of discrete variables. Next, the discrete variables are rounded-off and fixed, and a continuous optimization is performed over the subset of continuous variables.

\medskip

Another straightforward sequential approach consists in performing successively a discrete and continuous optimization process \cite{yan2006hybrid}. In \cite{gao2014hybrid}, a two-phase sequential approach is proposed: Sequential(ACO(Discrete, Partial), DE(~Continuous, Partial)). First, ACO performs a search in the discrete decision space $F_y$. A set of elite solutions are archived. Once the discrete values are fixed, the second phase applies a DE to optimize the continuous variables in the subspace $F_x$. Another complementary sequential methodology, Sequential(Meta(Discrete, Complete), Meta(Continuous, Partial)), has been investigated in \cite{stelmack1997concurrent}\cite{rao2005hybrid}. It consists first in discretizing the continuous variables with some specified resolution, and the problem is globally solved as a DOP, using for instance SA \cite{stelmack1998genetic} and GA \cite{rao2005hybrid}. Subsequently, the discrete variables are frozen, and a gradient is applied over the continuous variables to solve the CnOP subproblem \cite{stelmack1998genetic}, or a global relaxation of the MVOP \cite{rao2005hybrid}. 

\medskip

% Memetic & Matheuristic
Memetic approaches have also been considered. In \cite{lin2013mixed}, a global approach based on continuous relaxation is first applied. Then, a local search algorithm, based on Nelder-Mead algorithm, is applied on the subset of continuous variables, by fixing the values of discrete variables. Matheuristics combining metaheuristics and mathematical programming can also be considered. In \cite{nema}, branch-and-bound (B\&B) is sequentially combined with PSO to solve MIPs. First, B\&B is applied to define a feasible initial solution for PSO. Whenever, an improvement of the global best solution is found by PSO, this solution is passed to B\&B as a starting solution.

\medskip

% Analysis
Decomposition-based sequential strategies allow the reuse of standard metaheuristics in solving the subproblems. However, they ignore the correlation between the different types of variables by evolving separately the continuous and the discrete variables. Hence, the characteristics of the applied metaheuristics, the consistency and compatibility of their search operators (e.g. variation operators) can be destroyed. Moreover, the convergence of such methodology can be prohibitively slow and needs many sequential iterations. Instead of using only the type of variables as a criterion of decomposition, some strategies use some knowledge related to the structure properties of a MVOP to carry out the partitioning of the decision variables \cite{hua2006effective}.

%******************************************************************
\subsubsection{Nested approaches for MVOPs}
%******************************************************************

\begin{figure}[htb]
	\begin{center}
		\leavevmode
		\includegraphics[width=2.5in]{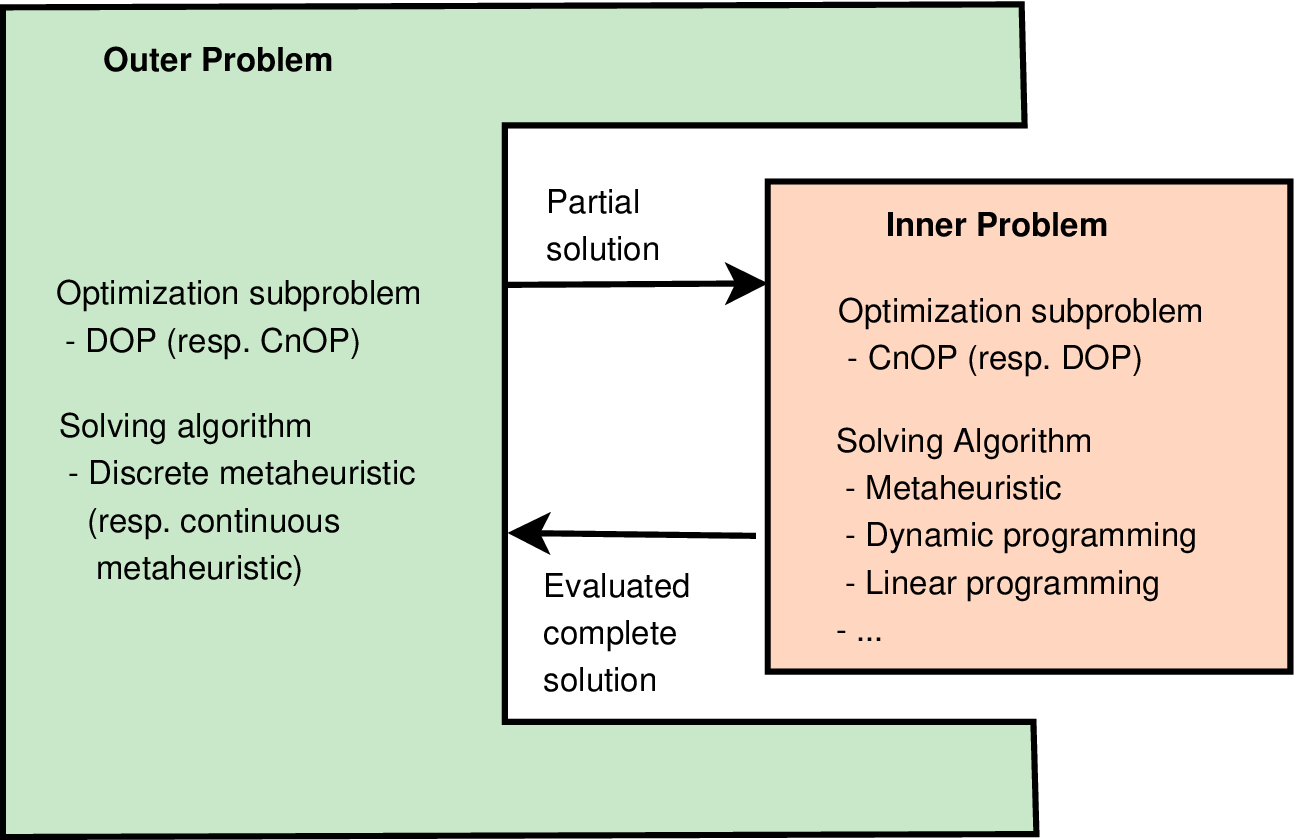}
		\caption{Nested decomposition-based approaches for handling MVOPs.}
		\label{fig:nested}
	\end{center}
\end{figure}

Nested approaches are generally based on a two-level decomposition strategy in which the entire domain of decision variables is partitioned into two levels, one involving the continuous variables and the other involving the discrete variables. The subproblems are solved in a hierarchical way, in which a subproblem associated to the continuous (resp. discrete) variables is optimized within the inner loop and the remaining discrete (resp. continuous) one at the outer loop level. Variables in one level are optimized for fixed values of the variable from the other level. Solutions at the outer subproblem can be evaluated once the inner problem is solved (Fig.\ref{fig:nested}). Hence, decision variables of a given partition are optimized independently for fixed values of the decision variables of the other partition \cite{praharaj1992two}\cite{lucidi2005algorithm}.

\medskip

Decomposition-based nested approach represents a major approach in mathematical programming: {\it bi-level programming} \cite{talbi2013taxonomy}\cite{talbi2013}\cite{sinha2017review}, {\it Danzig-Wolfe decomposition} for MILP (Mixed Integer Linear Programming) \cite{vanderbeck2006generic}, the {\it generalized Benders' decomposition}  \cite{geoffrion1972generalized} and the {\it outer approximation method} \cite{duran1986outer} for MINLP (Mixed Integer Non-Linear Programming) \cite{floudas1995nonlinear}. In general, nested mathematical programming methodologies have to solve relaxed problems in which the integer constraints are dropped, or they solve a serie of non-linear optimization problems in which some integer values are fixed. These methodologies are complex and they require the objective functions and constraints to be differentiable which restricts their application to many real-life problems \cite{cheung1997coupling}.

\medskip

The nested approach on the discrete variables $y$ at the outer level can be formulated as follows:
\begin{equation}
\label{nested-vxop}
\begin{aligned}
\text{min } f(x^*,y) & \\
\text{w.r.t } y \in F_y & \\
\text{s.t. } g(x^*,y) \leq 0 & \\
& \text{with } x^* = \text{argmin } f(x,y) \\
& \text{w.r.t } x \in F_x \\
& \text{s.t. } g(x,y^*) \leq 0 
\end{aligned}
\end{equation}

% Meta examples
Various algorithms in solving CnOPs and DOPs can be carried out to properly handle both optimization levels:
\begin{itemize}
	% Meta & Meta
	\item{\bf Combining metaheuristics:} a straightforward approach consists in applying metaheuristics (e.g. single or multiple) at both levels. Hence, a discrete (resp. continuous) metaheuristic deals with the discrete (resp. continuous) outer problem and another inner-level metaheuristic focuses on the continous (resp. discrete) one. The notation {\it Nested(Outer-Algo(Outer-Problem-Type, Inner-Algo(Inner-Problem-Type))} specifies the type and optimization algorithms used at the two levels. Some nested approaches have been proposed in the literature such as Nested(PSO(Discrete,GA(Continuous))) \cite{chanthasuwannasin2017mixed}, Nested(EA(Discrete,Gradient(Continuous))) \cite{roy2019mixed}, and Nested(ACO(Discrete,Gradient-based(Continuous))) \cite{wang2011simultaneous}. In \cite{lin2001co}, the same metaheuristic, an EA, is used at both levels. In \cite{roy2019mixed}, the algorithm first separates the discrete and continuous variables of the MVOP. An outer subproblem is solved using EGO (Evolutionary Global optimization), a surrogate-based evolutionary approach to solve expensive optimization problems \cite{jones1998efficient}. Considering the discrete variables fixed, a gradient-based optimizer is applied for inner CnOPs. It may be possible that for some initial discrete solutions, the continuous optimization may not find a feasible solution.
	
	% Meta & MP (LP, DP)
	\item {\bf Combining metaheuristics and mathematical programming:} a popular and obvious approach for solving MINLP (Mixed Integer Non-Linear Programming) problems defines the outer problem (i.e. master problem) by the discrete variables. Discrete metaheuristic approaches are then used to solve the master problem (e.g. GA \cite{garroussi2020matheuristic}\cite{chiam2019hierarchical}, DE \cite{balamurugan2008hybrid}, Hyper-heuristic \cite{gonzalez2022hyper}, PSO \cite{ye2019pso}). The inner problem includes only continuous variables. If the inner problem can be reduced to a linear programming (LP) problem, an exact linear solver can be applied \cite{gonzalez2022hyper}\cite{garroussi2017hybrid}\cite{chiam2019hierarchical}\cite{ye2019pso}. The same exact methodology has been used within an inner discrete subproblem solved by dynamic programming (DP), and a master continuous problem solved by DE (i.e. Nested(DE(Continuous,DP(Discrete)))) \cite{balamurugan2008hybrid}.
\end{itemize}

\medskip

% More complex hierarchical structure
The hierarchical decomposition of the variables could be based on other criteria than the type of variables. In \cite{zheng2013cooperative}, the problem is hierachically decomposed into many low-dimension subproblems by using some problem-specific knowledge, and PSO is used to solve each sub-problem.
% A relire \cite{lin2001co}: Nested - EA for the master discrete. EA for the continuous inner. The key ingredients of the algorithm consist of an integer-valued variable evolution and a real-valued variable co-evolution, so that the algorithm can be used to solve MINLP problems. the algorithm combines a local search heuristic (called acceleration) and a widespread search heuristic (called migration) to promote the search for a global optimum. coded as a real value in HDE The approach decomposes MINLP problems into a set of NLP sub-problems, and then applies HDE to evolve each sub-problem. The overall model (23) can be considered as an evolutionary combinatorial problem together with co-evolutionary multiple NLP sub-problems. In the outer loop, an integer- valued evolutionary algorithm performs a combinatorial optimization. In the inner loop, a real-valued co-evolutionary algorithm fulfills the minimization.

\medskip

% Analysis of Nested optimization
Nested decomposition-based optimization allows the reuse of efficient metaheuristics for solving CnOPs and DOPs subproblems, but ignores the correlation between the continuous and discrete variables and interrupts the co-evolution of variables. Indeed, as the interdependence between the variables is not explicitly carried out, this approach is predisposed to converge towards sub-optimal and non feasible solutions. Moreover, this approach often leads to an important number of function evaluations. Hence, an efficient inner metaheuristic must be used. In \cite{cheung1997coupling}, while the outer discrete problem is solved by a GA, an efficient modified grid search is applied to the continuous inner problem to speedup the search. An interesting idea to reduce the complexity of solving inner problems is to avoid making further iterations if the quality of the current solution remain far to the best found solution.

%******************************************************************
\subsubsection{Co-evolutionary approaches for MVOPs}
%******************************************************************

Co-evolutionary algorithms have been successfully applied for various optimization problems such as CnOPs, DOPs \cite{ma2018survey}, and bi-level problems \cite{legillon2012cobra}. Recently, some algorithms have been investigated for MVOPs. Co-evolutionary algorithms decompose a MVOP into multiple subproblems, and each subproblem is solved in a parallel way using a separate search process (e.g. generally an EA) to generate a partial solution \cite{potter1994cooperative}. Those multiple search processes cooperate to construct a complete solution for the problem by assembling the partial solutions found. 

\medskip

The main design questions for co-evolutionary approaches are (Fig.\ref{fig:coevolution}):
\begin{itemize}
	% Decomposition strategy
	\item {\bf Decomposition strategies}: an important issue in co-evolutionary approaches is the decomposition of the mixed vector of decision variables. A straightforward decomposition strategy of MVOPS consists in partitioning the decision space according to the type of variables (i.e. continuous and discrete subspaces). Constraint handling can also play a role into the decomposition strategy. In \cite{hiremath2012designing}, a constrained MVOP is transformed into an unconstrained problem using augmented Lagrangian and then formulating the MVOP within a Min-Max formulation using additional decision variables (i.e. Lagrangian multipliers).
	
	\begin{figure*}
	\centering
			\includegraphics[width=0.9\textwidth]{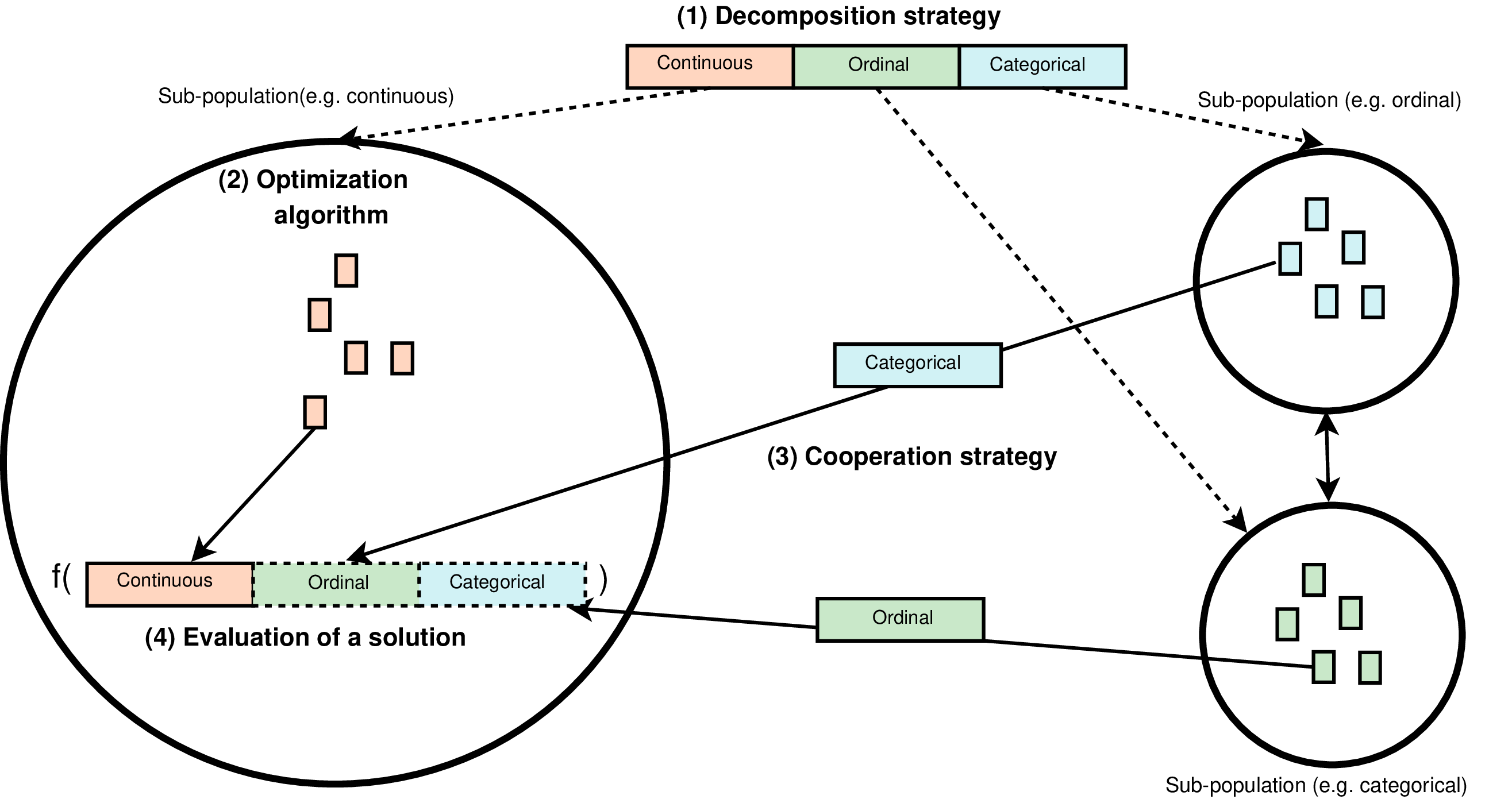}
			\caption{Decomposition-based co-evolutionary approaches for handling MVOPs.}
			\label{fig:coevolution}
	\end{figure*}
	
	\medskip
	
	% Variable interaction
	To minimize the relationships between the subproblems, the decomposition should also be based on the variables interaction \cite{omidvar2013cooperative}. Indeed, if the interacting decision variables are not in the same subproblem, the search tends to converge towards local optima \cite{van2004cooperative}. Many partitioning strategies based on variable interaction may be designed: 
	\begin{itemize}
		% Static/dynamic/adaptive
		\item {\bf Static versus dynamic decomposition}: most of the proposed decomposition strategies for MVOPs are static. In static strategies, a priori decomposition of the variables is carried out and is fixed during the search, while in dynamic strategies the decomposition of the mixed decision space may be updated during the search. Dynamic decomposition allows to define the number of subproblems and their associated variables in an adaptive way, by learning the extracted knowledge related to variable interactions \cite{mei2016competitive}.
		
		% Exclusive versus overlapped decomposition
		\item {\bf Disjoint versus overlapped decomposition}: in general, each decision variable belongs to a single subproblem. Overlapping the variables into subproblems can be an interesting strategy if multiple interactions between the variables occur \cite{strasser2016factored}. A complete overlapping of variables has been used in \cite{vinko2008global}, where various metaheuristics (i.e. DE, PSO, GA, and SA) are cooperating in parallel to solve the same VMOP.
		
		% Knowledge-based
		\item {\bf Knowledge-based decomposition}: some proposed approaches are application-specific and include some knowledge on the MVOP to decompose the MVOP \cite{pan2020effective}. In this case, the application expert uses some knowledge on the variable intraction. For instance, in complex engineering design problems, the well known correlation between the design variables are used to generate manually the subproblems \cite{zhang2013hybrid}. Some other MVOPs involve the joint optimization of various optimization problems. In \cite{yuan2020co}, the MVOP involves two problems namely group scheduling and job scheduling within each group.
	\end{itemize}
	
	% \cite{zhang2013hybrid}: Among the limited number of related efforts that combine them, most follow a sequential decomposition strategy. This sequential strategy has been successful in addressing the combined large-scale problem but the approach does not capture the coupling that exists between the aircraft design and airline allocation disciplines.
	
	% Metaheuristic algorithm
	\item {\bf Optimization algorithm}: a metaheuristic is associated to solve a given sub-problem (i.e. subset of decision variables). Each metaheuristic applies variation operators to each sub-problem to generate new solutions. The most popular metaheuristic algorithms used are based on EAs (e.g. GA \cite{zhang2013hybrid}\cite{yuan2020co}). Some proposed co-evolutionary approaches use heterogeneous metaheuristics. In \cite{hiremath2012designing}, two different metaheuristics, namely PSO and artificial immune systems (AIS), are used to solve the subproblems.
	
	% Cooperation and replacement
	\item {\bf Cooperation strategies}: in addition to the decomposition strategy, the cooperation between the subproblems is a key issue \cite{shi2017reference}. The evaluation of a solution of each subproblem requires the cooperation between the subproblems. The selection of solution(s) to be communicated can be either solution-centric or population-centric \cite{popovici2012coevolutionary}. Most of the proposed strategies proposed in the literature select a single solution. Many strategies can be used such as the best solution \cite{yang2008large}, worst solution \cite{wiegand2001empirical}, random solution \cite{de2016cooperative}, roulette/tournament selection \cite{son2004hybrid}, and elite solutions \cite{glorieux2015improved}. 
	\medskip
	
	% solution evaluation
	\item {\bf Evaluation of solutions}: the evaluation of a subroblem solution is approximated based on how well the solution cooperates with other subproblems to generate a complete solution. In general, a solution in a given subproblem is evaluated by using a single solution from each subproblem. For instance, in \cite{hiremath2012designing}, each metaheuristic communicates the {\it best solution} found according to its subproblem. Then, the $jth$ solution $U_{i,j}$ of the $ith$ subpopulation is evaluated as following: $f(V_1, . . . , V_{i-1},U_{i,j}, V_{i+1}, . . . , V_n)$ where $V_i$ is the best solution from the $ith$ subpopulation. 
	
	\medskip
	
	Instead of constructing a single complete solution to evaluate, some evaluation strategies construct multiple complete solutions to assess the evaluation of solutions into a given subproblem. The most representative strategies are the {\it complete cooperative selection} and the {\it archive-based cooperative selection}. In the complete cooperative selection strategy, each solution in a subproblem is evaluated by assembling it with all candidate combinations from other subproblems \cite{panait2006selecting}, while in archive-based cooperative selection, solutions are selected from a given archive (e.g. non-dominated solutions \cite{bucci2005identifying}). Then, those multiple evaluations are used into a Pareto dominance framework (e.g. quality, diversity) \cite{bucci2005identifying} or agregated to estimate the quality of the solution (e.g. best, worse, average) \cite{wiegand2001empirical}.
\end{itemize}

\medskip

% Advantages & Inconvenient
Co-evolutionary algorithms have many advantages. Parallel solving of sub-problems allows to speedup the search and solve large scale MVOPs. It encourages to maintain the diversity of solutions. It allows also to improve the robustness and adaptability in solving dynamic MVOPs \cite{nguyen2007analysis}. However, co-evolutionary algorithms can be trapped by local optima. Indeed, the speed of convergence and state of each sub-problem may be inconsistent \cite{song2020variable}. The subproblems may be different in size and complexity. The allocation of computational resources to the subproblems is also a crucial issue in the design of efficient co-evolutionary algorithms. There is a possibility that subproblems with small search space size or complexity converge first or trapped by local optima, while subproblems with large search space or higher complexity require more iterations to converge towards good quality solutions. Some machine learning strategies (e.g. Q-learning) can be applied to adapt the size or the budget (e.g. population size) of each subproblem \cite{song2020variable}\cite{abdelkhalik2012dynamic}. 

\medskip

% Multiple approaches
According to the proposed taxonomy, combining multiple approaches such as global and decomposition-based approaches can be performed. Some methodologies based on multiple approaches have been investigated in the literature to solve MVOPs. In \cite{peng2021multi}, a hybrid approach combining a global approach (i.e. relaxation with DE) and a decomposition-based approach (i.e. sequential) is proposed. The combination between global and decomposition-based nested approaches has been investigated in \cite{hansen2008multilevel}\cite{venter2004multidisciplinary}\cite{stelmack1998genetic}, by using various metaheuristics such as ES, PSO and gradient algorithms.

%**************************************************************
\section{A taxonomy of metaheuristics for VMVOPs}
%**************************************************************
\label{mvmop}

% Examples of applications
Many real-world applications are formulated as variable-size MVOPs (VMVOP) such as multidisciplinary design optimization (MDO) \cite{sobieszczanski1997multidisciplinary}, classification  \cite{bandyopadhyay2001pixel}\cite{maulik2009modified}, grouping problems \cite{falkenauer1994new}. MDO involves mutually dependent disciplines that must synchronized. For instance, in the design of an aircraft wing, the following variables are involved: the number of spars in the wing (integer), their location (continuous), and the type of material used (dimensional). Moreover, many new problems in automated deep machine learning (AutoDeepML) such as neural architecture search (NAS) and hyperparameter optimization represent an important family of VMVOPs \cite{talbi2021automated}\cite{costa2018hierarchical}. Variable-length optimization problems (VOPs) in which there is a single type of variables is a sub-family of MVMOPs. 

\medskip

\begin{figure*}[t!]
	\centering
		\includegraphics[width=0.95\textwidth]{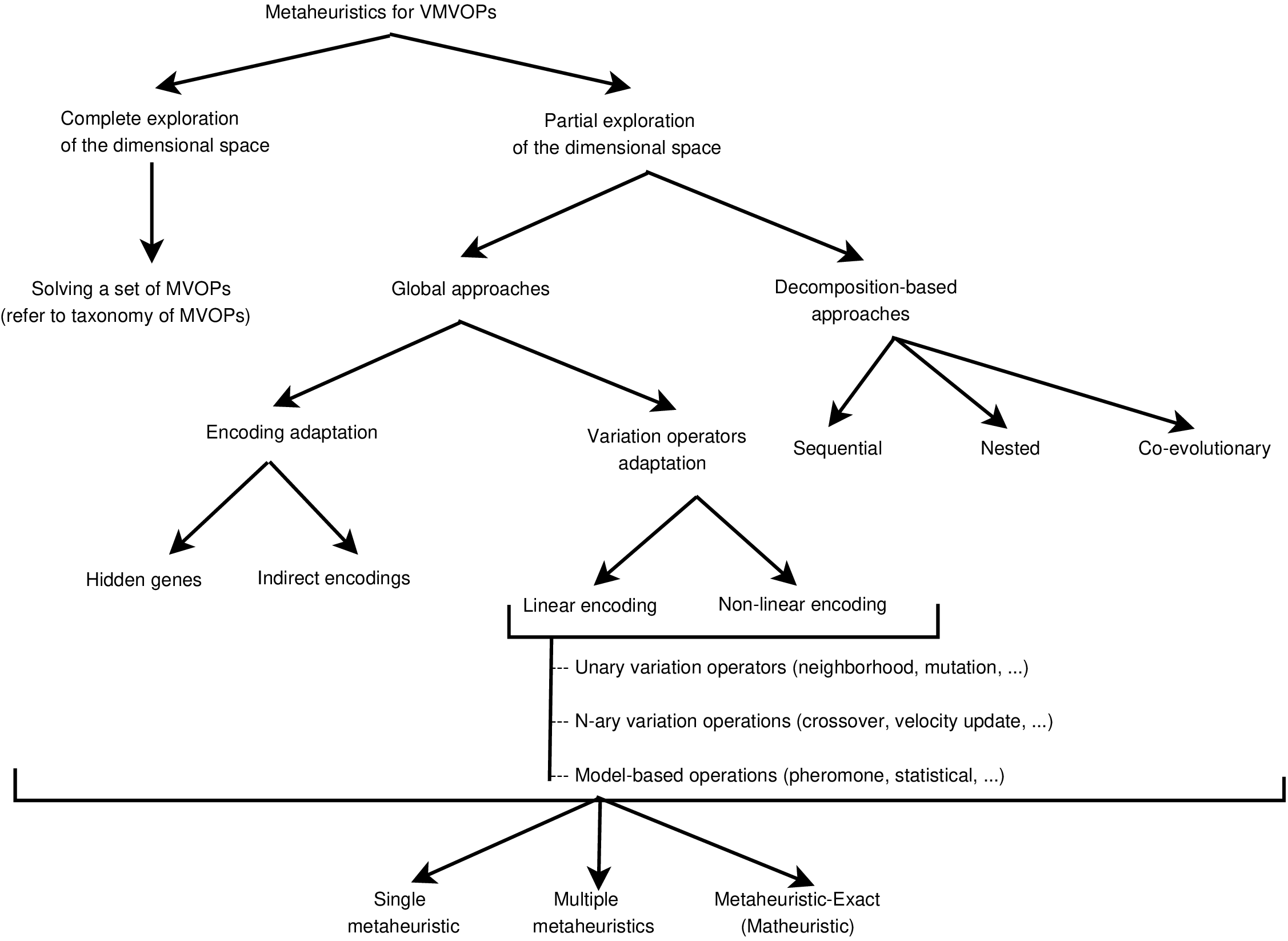}
		\caption{A taxonomy of metaheuristic methodologies to solve VMVOPs.}
		\label{fig:taxonomyVMVOP}
\end{figure*}

% Complete exploration - Transformation to a VMOP - Complete search over the dimensional space
If the size of the dimensional space is reasonable, its {\it complete exploration} can be performed (Fig.\ref{fig:taxonomyVMVOP}). Hence, the VMVOP will be transformed into a set of independent MVOPs. Indeed, for every solution of the dimensional space, one can generate a fixed-size MVOP:
\begin{equation}
\begin{aligned}
\text{for all } & z_i \in \mathcal{F}_z = \{z_1,z_2,...,z_{n_z}  \} & \\
& min \hspace{0,2cm} f(x,y,z_i), f: \mathcal{F}_x(z_i) \times \mathcal{F}_y(z_i) \longrightarrow \mathbb{R} & \\   
& \text{s.t. } x \in \mathcal{F}_x(z_i) \subseteq  \mathbb{R}^{n_x}, y \in \mathcal{F}_y(z_i) \subseteq Z^{n_y} & \\ 
\end{aligned}
\end{equation}

% The complete nested approach methodology can be formulated as follows:
%\begin{equation} \label{complete}
%\begin{aligned}
%min \hspace{0,2cm} f(x,z_p,w_q) \hspace{0,2cm} & \forall w_q  \in F_w, \forall z_p \in F_z(w_q)  \\
%w.r.t. \hspace{0,2cm} &x \in F_x(w_q)  \\
%&z \in F_z \prod_{d=1}^{n_z(w)}{F_{z_d}}  \\
%subject to  & g(x,z_p,w_q) \leq 0 \\
%\end{aligned}
%\end{equation}

% Analysis
In that case, any metaheuristic approach for MVOPs can be applied for solving this set of MVOPs. Even though this complete methodology is massively parallel, it still prohibitive in terms of computational time if the size of the dimensional space is not small. Indeed, this approach is unfeasible for VMVOPs characterized by a large dimensional space and/or expensive VMVOPs where a complete enumeration of the dimensional space is computationally intractable. 

\medskip 

% Examples
This approach has been used in engineering design problems such as launcher design \cite{frank2016evolutionary}. Once all MVOPs are solved by a GA, all the results are agregated to generate the best found solutions. Non feasibility of some dimensional variables values (e.g. incompatibility between dimensional choices) allows to reduce the number of MVOPs to solve. Human expertise and application specific considerations can also be used to reduce the dimensional space. The computational budget (e.g. number of iterations of the metaheuristic, number of objective function evaluations) assigned to solve each MVOP is an important issue. For instance, it is interesting to assign more computational budgets to more promising MVOPs. In the following, we will extend the taxonomy for VMVOPs in the case where a {\it partial exploration} of the dimensional space is applied (Fig.\ref{fig:taxonomyVMVOP}).

%******************************************************************
\subsection{Global approaches for VMVOPs}
%******************************************************************

In global approaches for VMVOPs, a number of additional challenges arize because of the presence of dimensional variables and the dynamically varying decision space.

%***************************************
\subsubsection{Encodings adaptation}
%***************************************

In VMVOPs, the natural encoding of solutions is defined as a variable-length vector of mixed variables. To reuse standard variation operators, some fixed-size representations have been proposed such as {\it hidden genes} and {\it indirect encodings}.

\medskip

\begin{figure*}[t!]
\centering
		\includegraphics[width=0.95\textwidth]{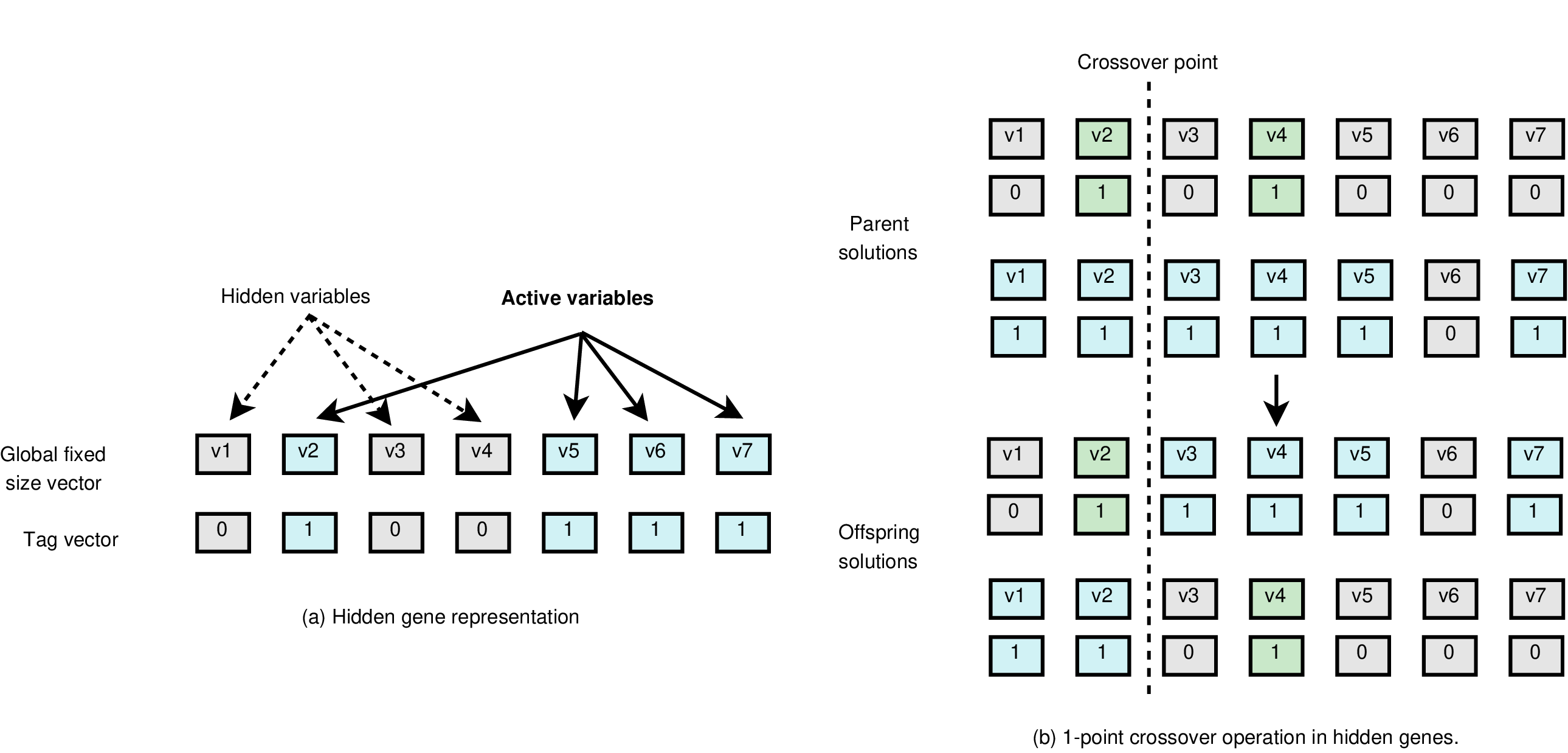}
		\caption{Hidden genes representation and variation operators for VMVOPs.}
		\label{fig:hiddengenes}
\end{figure*}

% Hidden genes
{\bf Hidden genes\footnote{Also called non coding segments.}:} this fixed-size encoding has been introduced in \cite{zebulum2000variable}\cite{gad2011hidden}. A solution is represented by two structures: a global mixed vector which encodes all mixed decision variables of a VMVOP, and a binary tag vector which specifies which variables are expressed (Fig.\ref{fig:hiddengenes}a). The global mixed vector contains the maximum number of variables that characterizes the VMVOP at hand. Not all decision variables are taken into account in the evaluation of a given solution. To each variable is associated a binary tag. If the tag is not activated, the corresponding decision variable will not participate to the evaluation of the solution \cite{abdelkhalik2013hidden}. For a given VMVOP, there exists various ways to define dimensional variables in order to describe the global and the tag vectors \cite{gamot2023}. This approach has been mainly used in EAs (e.g. GA \cite{abdelkhalik2012dynamic}, DE \cite{abdelkhalik2013autonomous}\cite{chen2015reconfiguration}) but can be naturally extended to other families of metaheuristics such as PSO \cite{mukhopadhyay2014identifying}. 

\medskip

% Variation operators
As the representation of solutions has a fixed length, the standard variation operators of metaheuristics can still be performed. All the elements of the two vectors, even the hidden ones, take part in the variation operators. Unary variation operators (e.g neighborhoods, mutation) will alter both the value of the mixed decision variables and the tag binary values. Crossover operators act in both vectors: global and tag vectors (Fig.\ref{fig:hiddengenes}b). It can also be applied separately \cite{darani2018space}. Statistical models in model-based metaheuristics (e.g. EDA, ACO) can also be easily constructed \cite{chen2015reconfiguration}.

\medskip

% Advantesges, inconvenients
This approach is simple and easy to implement. However, the variation operators can generate inconsistent and non feasible solutions, which result in additional computational time and ineffective numerical operations. Moreover, for VMVOPs containing a large number of dimensional variables, the global mixed vector might become substantially large.

\medskip

% Indirect encodings
{\bf Indirect encodings:} indirect fixed-size encodings (e.g. permutations, matrices) represent partial implicit representations of variable-size mixed solutions \cite{rothlauf2006representations}. A problem-specific decoding procedure must be developed to map the partial implicit fixed-size representation to a complete explicit variable-size representation. Indirect encodings may define a many-to-one mapping, in the sense that many indirect encodings may represent the same solution. Once the fixed-size indirect encodings defined, standard variation operators can be applied. For instance, this approach has been integrated in ACO \cite{gao2021adaptive}, EA \cite{gao2021hybrid}\cite{reuter1997heterogeneous}, DE \cite{wang2018hybrid}, and memetic algorithms \cite{gao2020memetic} to solve various VMVOPs. 

%***************************************
\subsection{Variation operators adaptation}
%***************************************

% Linear versus non-linear
Variable-length encodings may be represented by {\it linear} or {\it non-linear representations}. In a linear representation, all variables are at the same level. It is well adapted for VOPs characterized by a single type of variables (e.g. all continuous, all binary, all discrete ordinal). Many papers in the evolutionary computation (resp. swarm intelligence) literature address this family of encodings which is called variable-length chromosomes (VLC) \cite{merlevede2019homology}\cite{ryerkerk2019survey} (resp. variable number of dimensions (VND) \cite{marek2022another}\cite{kiranyaz2009fractional}). Standard variation operators can be extended easily for such encodings, in case the position independent encoding of variables is ensured in the linear representation.

\medskip

% variation operators for variable length linear encodings (VOPs)
Mutation including operators such as adding and deleting variables has been introduced in \cite{neumann2012targeted}. Adapted crossover operators have been proposed in \cite{chen2015reconfiguration}\cite{zebulum2000variable}\cite{dwivedi2018learning}. For instance, various crossover operators preserving high-quality buidings blocks in variable-length linear representations have been proposed in \cite{ryerkerk2012optimization}\cite{hutt2007synapsing}\cite{goldberg1989messy}\cite{harvey1992saga}\cite{burke1998putting}\cite{dasgupta1992nonstationary}. In \cite{hutt2007synapsing}, the authors introduce a synapsing variable-length crossover (SVLC). In the SVLC crossover, the common components of the parents are preserved in the offsprings, and the different components are exchanged or removed, independent of the length of such differences. Based on the common sequence similarity between the parents, the operators select non random crossover points. Movement update operations (e.g. PSO) have also been adapted to variable-length chromosomes \cite{yan2009density}\cite{yangyang2004particle}\cite{kadlec2018particle}\cite{marek2022another}. In \cite{tran2018variable}, the authors extend the PSO movement updating mechanism proposed in the comprehensive learning PSO (CLPSO) \cite{liang2006comprehensive}, which may change the length of a particle.

\medskip

\begin{figure*}[t!]
\centering
		\includegraphics[width=0.8\textwidth]{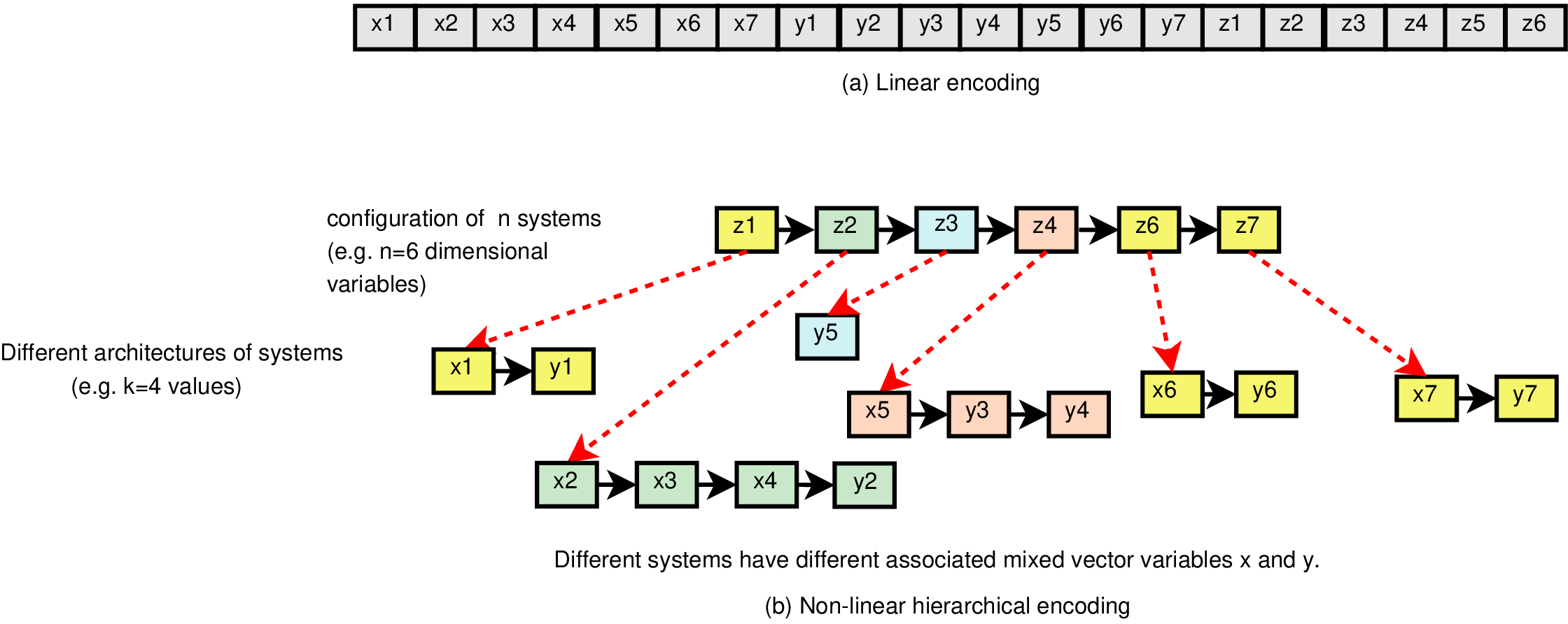}
		\caption{Linear and non-linear representations for a simple VMVOPs.}
		\label{fig:hierarchy}
\end{figure*}

A non-linear encoding based on hierarchical linked list of variables is more adapted to VMVOPs \cite{talbi2006hierarchical}\cite{nyew2015variable}\cite{liu2019coordinated}\cite{gentile2019structured}. The variables are classified as dependent or independent variables and are linked by both vicinity and hierarchical relations. Let us illustrate those encodings with a simple design problem, in which we have to configure a set of $n$ systems. Each system is characterized by a dimensional variable of order $k$. Each given configuration of the system is defined by a variable set of mixed variables. In the linear encoding, the variables are stacked in a linear vector of variable-length (Fig.\ref{fig:hierarchy}). The data structure does not link explicitly the dependent variables. In the non-linear encoding, the variables are hierarchically structured in different levels. Solid arrows connect next neighbor variables, and dotted arrows connect child variables (Fig.\ref{fig:hierarchy}). For instance, the value of the dimensional variable $z_1$ will define the number and type of variables in the second level (e.g. $x_1$ and $y_1$). Variables that are pointed are dependent variables. Their existence depends on the value of the ancestor variables. Hence, the nodes $z_1$, $x_1$ and $y_1$ are dependent nodes. 

\medskip

Standard variation operators typically operate on fixed-size linear encodings. New variations operators have to be designed for hierarchical representations:
\begin{itemize}
	% Unary operators
	\item {\bf Unary operators (e.g. neighborhood, mutation)}: for a given solution, many neighborhoods and mutation operators may be defined. The following design questions arize:
	\begin{itemize}
		% Number of neighborhoods
		\item {\bf Definition of neighborhoods and mutations operators}: the number of neighborhoods and mutation operators is related to the number and type of variables associated to the VMVOP. Those operators perform at all levels of the hierarchy. However, the semantic correctness according to the new value of the variable must be ensured. Unary operators for non dimensional variables will not affect the structure of the graph modeling the hierarchical encoding. Indeed, the new value must be consistent with the type of variable (e.g. continuous, discrete categorical, discrete ordinal) \cite{liu2019coordinated}. The operators associated to dimensional variables (i.e. variables with dependent childs) have an impact on the number and type of the other variables. The whole subtree (i.e. size, type of variables) of dimensional variables is transformed according to its new value (Fig.\ref{fig:hierarchical-mutation}). Moreover, changing a value of a given dimensional variable has to fix the values of the new dependent child variables (Fig.\ref{fig:hierarchical-mutation}). One can randomly initialize those values, or search the sub-optimal (resp. optimal) values using some heuristics (resp. exact algorithms).
		% Examples: 
		% \cite{liu2019coordinated}: Differential Evolution Algorithm. Hierarchical mutation. 
		
		% Selection and order of exploration of neighborhoods
		\item {\bf Order of application of the various operators}: once the unary operators defined, the order of application of those operators is a crucial issue. Operations at the higher hierarchies of the tree generate greater disruption of the solution (i.e. lower degree of locality, higher diversification) than those on the low levels of the hierarchy (i.e. higher degree of locality, higher intensification). A higher degree of locality means that small perturbations to the representation (i.e. genotype) result in small perturbations to the solution (i.e. phenotype).
	\end{itemize}
	
	\begin{figure}[t!]
	\centering
			\includegraphics[width=2.5in]{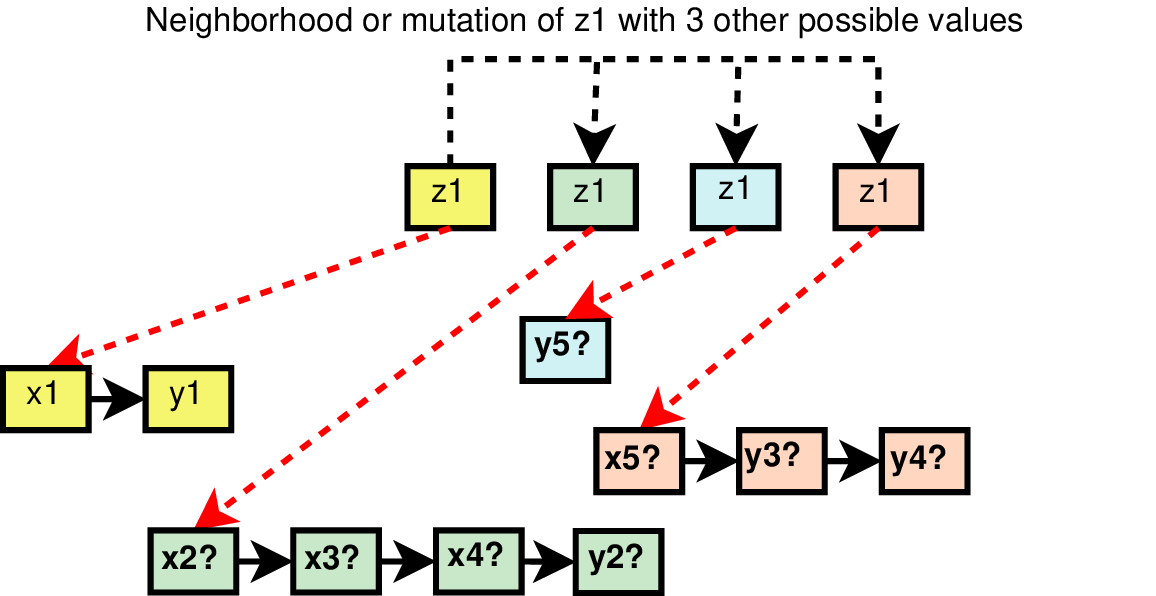}
			\caption{Unary operator (e.g. neighborhood, mutation) of a dimensional variable in hierarchical encodings.}
			\label{fig:hierarchical-mutation}
	\end{figure}
	
	% N-ary operators
	\item {\bf N-ary operators (e.g. crossover, particle update)}: as for unary operators, the N-ary operators can be applied at different levels of the hierarchy. For instance, standard crossover operators (e.g. 1-point, n-point, uniform) can be applied at the first layer of the hierarchical encoding \cite{talbi2006hierarchical} (Fig.\ref{fig:hierarchical-crossover}). However, all the subtrees (i.e. dependent variables) must be inherited from the parents. Inheriting only a part of the subtree may result in non feasible or inconsistent solutions. More complex problem-specific crossover operators can be designed \cite{nyew2015variable}. The adaptation of the movement update operators (e.g. PSO) hierarchical representations is a more complex operation to define, and has not been widely studied in the literature.
	
	\begin{figure*}[t!]
	\centering
			\includegraphics[width=0.8\textwidth]{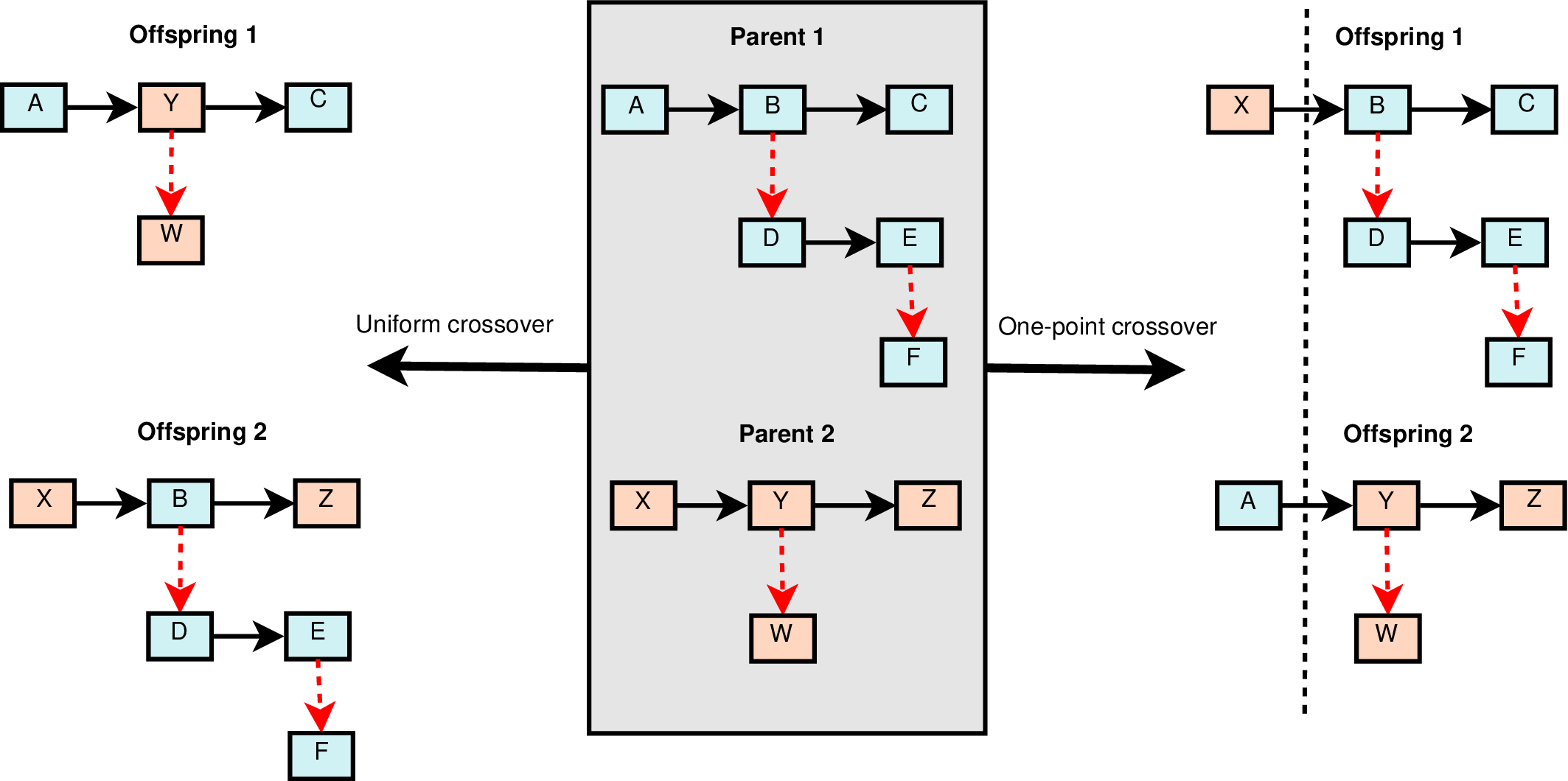}
			\caption{Crossover operations in hierarchical encodings.}
			\label{fig:hierarchical-crossover}
	\end{figure*}
	
	\medskip
	
	% Model construction
	{\bf Model-based operators (e.g. pheromone in ACO, statistical model in EDA):} there has been growing interest in adapting EDAs to construct statistical models on tree representations. In \cite{shan2006survey}, the authors extend EDA to genetic programming (GP) tree representation. GP evolves tree structures which encode computer programs (i.e. mathematical function) \cite{koza1992programming}. The tree has a limited hierarchical structure in which the terminal set could be $\{ x, y, z \}$ and the non-terminal set could be operations such as $ \{ +,-,×,/, sin, cos, log, exp \} $. Very few works in the literature address those important scientific challenges.
	
\end{itemize} 

% Avantages inconventients
Compared to the linear encodings, the hierarchical encoding is more adapted to VMVOPs.  It allows more flexibility in the design of efficient variation operators, which generate valid solutions. Compared to the hidden genes approach, hierarchical encoding has the advantage of not performing computationally wasteful variation operations. However, the encoding is much more complex and often requires problem-specific knowledge.

%*******************************************************************************************************************************************************
%*******************************************************************************************************************************************************

%**************************************************************
\subsection{Decomposition-based approaches for VMVOPs}
%**************************************************************

The main issue in decomposition-based approaches to solve VMVOPs is the handling of dimensional variables which induce variable-size vectors of mixed variables. 

%******************************************************************
\subsubsection{Sequential approaches for VMVOPs}
%*******************************************************************

The sequential approach presented in section~\ref{sequential} may be easily extended to VMVOPs. The design questions to be adapted are: 
\begin{itemize}
	% Sequence of optimization
	\item {\bf Sequence of optimization}: one can alternate iterations in which the dimensional variables are first considered fixed, with iterations in which the continuous and discrete variables (i.e. MVOP) are optimized (Fig.\ref{fig:sequentialVMOP}). Indeed, once the dimensional variables are fixed, a VMOP has to be solved \cite{isebor2009constrained}. This sequential approach can be defined as follows:
	\begin{equation}
	\begin{aligned}
	\text{min } f(x,y,z^*) & \longrightarrow^{(x^*,y^*)}  & \text{min } f(x^*,y^*,z)   \\
	\text{w.r.t } x \in F_x(z^*), y \in F_y(z^*)  &   & \text{w.r.t. } z \in F_z \\
	\text{s.t. } g(x,y,z^*) \leq 0  & \longleftarrow^{z^*}  & \text{s.t } g(x^*,y^*,z) \leq 0
	\end{aligned}
	\label{sequential-vxop}
	\end{equation}
	
	In \cite{audet2001pattern}, a sequential approach applying continuous search (e.g. pattern local search algorithm) followed by discrete search is proposed. 
	
	\begin{figure*}[htb]
	\centering
			\includegraphics[width=0.9\textwidth]{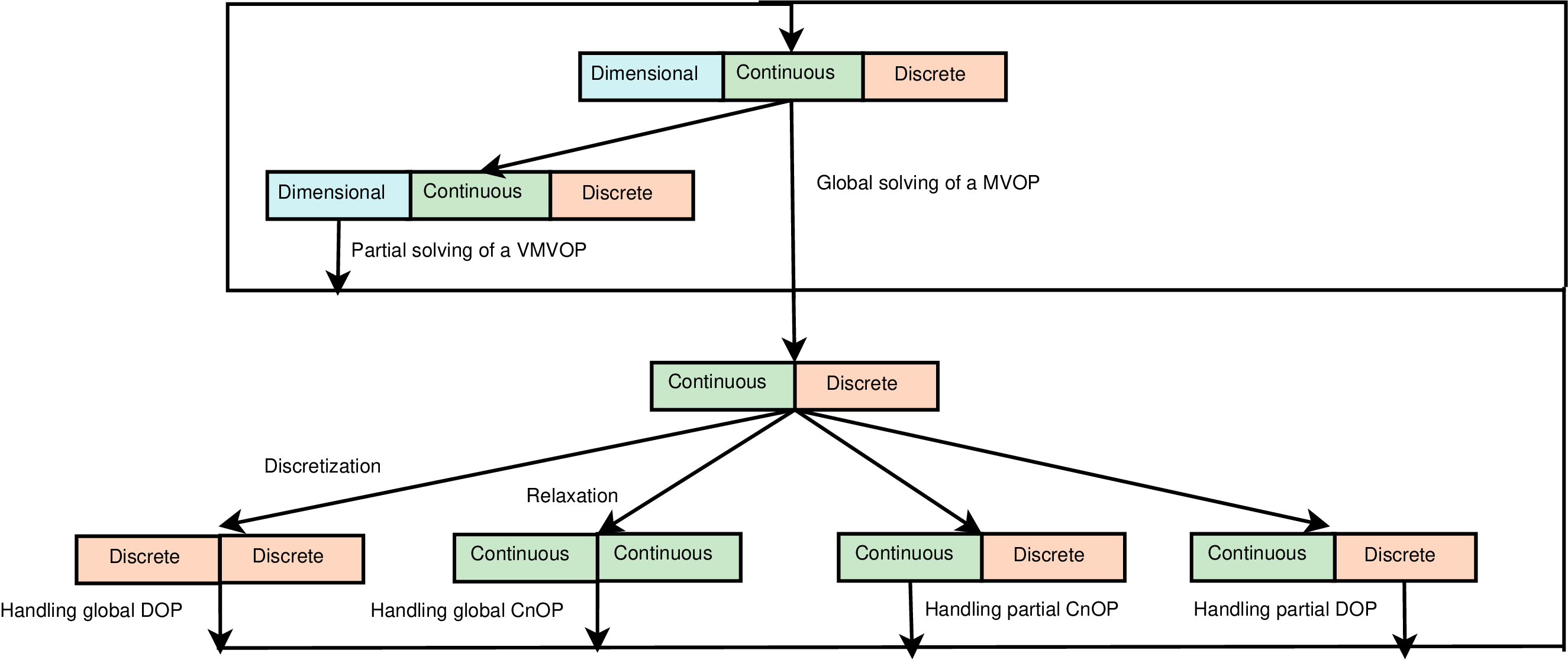}
			\caption{Decomposition-based sequential strategies for solving VMVOPs.}
			\label{fig:sequentialVMOP}
	\end{figure*}
	
	% Complete or partial optimization
	\item {\bf Complete versus partial optimization}: partial optimization allows to reduce the size of decision space, which is suitable for large scale VMVOPs (Fig.\ref{fig:sequentialVMOP}). In \cite{kim2005variable}, a multi-stage partial GA strategy is applied in which the design space is gradually refined during the search. Maintaining the diversity in the population in such an approach is a very important issue. The partial optimization can be based on the type of variables or on some problem-specific knowledge. In \cite{venter2004multidisciplinary}, a problem-specific knowledge is used to carry out a partial sequential optimization. At each phase, a VMOP mixing continuous and discrete optimization has to be solved.
\end{itemize}

% Bi-objective approach \cite{liu2021biobjective}.
% Pattern search (reference de base) \cite{torczon1997convergence}.

% Examples

% Similarly to the nested approach, a more general formulation can be defined in case both iterations of the sequential algorithm are tasked with optimizing part of the continuous and discrete variables. Notable examples of the sequential approach to VSDSP are the mesh-based optimization algorithms for mixed-variable problems presented in \cite{abramson2004filter}\cite{abramson2009mesh}\cite{audet2001pattern}\cite{kokkolaras2001mixed}\cite{lucidi2005algorithm}.

%For instance, given the current feasible solution $(x_k,y_k,z_k)$ and fixing the discrete variables $(y_k, z_k)$, the subproblem associated to the continuous variables can be defined as:
%\begin{align}
%min f(x,y_k,z_k,w_k) \\
%x \in \mathbb{F}_x(y_k,z_k,w_k)
%\end{align}

% Neighborhood in single-solution based algorithms - Exploration of the neighborhood

% Mesh: global neighborhood
% A mesh is a discrete set of points. The mesh is the direct product of the union of a finite number of lattices in $\mathbb{R}^{n_x}$ with the integer space $\mathbb{Z}^{n_d}$.

\medskip

\subsubsection{Nested approach for VMVOPs}
%**************************************************************

\begin{figure*}[t!]
\centering
		\includegraphics[width=0.9\textwidth]{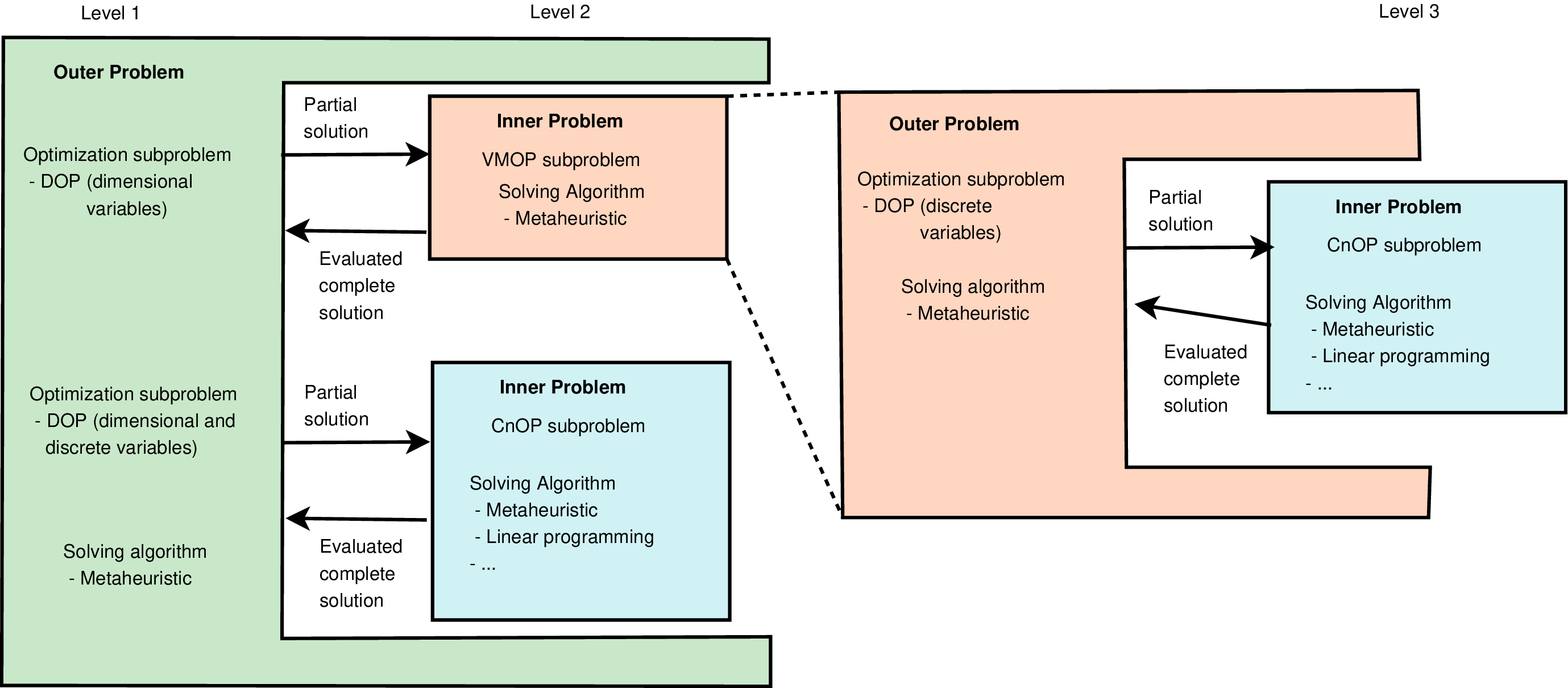}
		\caption{Nested aporoaches for VMVOPs.}
		\label{fig:nestedVMOP}
\end{figure*}

% Partitioning - Dimensional variables - Complete or heuristic
A straightforward nested approach consists in optimizing at the outer-level (resp. inner-level) the dimensional variables (resp. continuous and discrete variables). A metaheuristic search over the dimensional variables can be applied (Fig.\ref{fig:nestedVMOP}). When fixing the dimensional variables at the outer-level, one has to solve a MVOP at the inner-level. Indeed, a MVOP subproblem is associated for each possible combination of dimensional variables. For each iteration of the outer loop, which specifies the values of the dimensional variables, a complete optimization of the problem with respect to the relevant continuous and discrete variables is performed in the inner loop. Then, any metaheuristic presented in section\ref{mvop} can be used to solve the VMOP. The nested approach on the dimensional can be formulated as follows:
\begin{equation}
\begin{aligned}
& \text{min } f(x^*,y^*,z) & \\
& \text{w.r.t } z \in F_z  & \\
& \text{with } (x^*,y^*) = & \text{argmin } f(x,y,z)  \\
& &  \text{w.r.t } x \in F_x(z), y \in F_y(z)  \\
& & \text{s.t } g(x^*,y^*,z) \leq 0  \\
\end{aligned}
\label{nested-vxop}
\end{equation}

This approach has been used in \cite{englander2012automated}\cite{englander2012automated}\cite{chilan2013automated}. In \cite{englander2012automated}, a binary GA algorithm (resp. hybrid PSO and DE) is applied at the outer-level (resp. inner-level). This approach is not efficient in case the inner-level MVOP is complex to solve, and there is a large proportion of inner-level MVOPs with infeasible solutions. Another two-level nested approachs maps the dimensional and the discrete variables at the outer-level and the continuous variables at the inner-level. Finally, one can also use a 3-level nested approach for handling respectively the dimensional, discrete and continuous variables (Fig.\ref{fig:nestedVMOP}).

%**************************************************************
\subsubsection {Co-evolutionary-based approaches for VMVOPs}
%**************************************************************

% Re-visit the design questions
To our knowledge, no co-evolutionary approaches have been applied to VMVOPs. Most of the proposed strategies have been employed to VOPs with single type of variables \cite{song2020variable}. One can extend the design question of co-evolutionary algorithms as following:
\begin{itemize}
	% Decomposition strategies
	\item{\bf Decomposition strategies}: promising decomposition strategies for VMVOPs can be applied in the dimensional space. Decomposition should also be based on the dimensional variables interaction. The decomposition criteria shown in section~\ref{mvop} can also be used here: static versus dynamic, disjoint versus overlapped, and knowledge-based decomposition. For instance, a static and disjoint partitioning of the dimensional space can be performed using a sequential or random division of the dimensional space \cite{song2020variable}. 
	
	% Optimization algorithms
	\item {\bf Optimization algorithm}: the selected metaheuristic depends on the type of the generated sub-problems (i.e. VMVOP or MVOP). If the generated subproblems are VMOPs, any metaheuristic shown in section\ref{mvop} can be used. In \cite{abdelkhalik2012dynamic}, the subpopulations have fixed-size design spaces and are randomly initialized. All the solutions in each sub-population have the same encoding length, and then global relaxation GAs are applied.
	
	\item {\bf Cooperation strategies and evaluation of solutions}: the same cooperative strategies as for MVOPs can be used. However, the evaluation of solutions represents a difficult challenge. Indeed, assembling the various partial solutions to construct a global consistent and feasible solution is complexified by the presence of dimemsional variables.
\end{itemize}

%*****************************************
\section {Conclusions and perspectives}
%*****************************************
\label{conclusion}

% Conclusion
This paper presents a unified taxonomy for metaheuristics in solving (V)MVOPs in an attempt to provide a common terminology and classification mechanisms. It provides a general mathematical formulation and concepts of (V)MVOPs, and identifies the various solving methodologies than can be applied. The advantages, the weaknesses and the limitations of the presented methodologies have been discussed. 

\medskip

% Perspectives
Nevertheless, developing efficient and effective metaheuristics for handling (V)MVOPs is still challenging. One has to extend the proposed taxonomy to include surrogate-based metaheuristics for (V)MVOPs. Very few works try to solve (V)MVOPs that involve both expensive objective functions and high number of mixed variables. Indeed, most of the work in the literature dealing with Bayesian optimization applies to medium-size continuous optimization problems \cite{morar2021bayesian}\cite{pelamatti2021mixed}\cite{manson2021mvmoo}. 

\medskip

% MOP, dynamic, bi-level, uncertainty
Extending the proposed taxonomy and critical analysis for solving multi-objective (V)MVOPs \cite{tanabe2020easy}\cite{li2019variable}, bi-level (V)MVOPs \cite{tang2016class}\cite{kleinert2021survey}, dynamic and uncertain (V)MVOPs \cite{belotti2013mixed}\cite{yao2011review} is a challenging task. Indeed, many real-life (V)MVOPs in science and industry involve multiple objectives, Stackelberg sequential games and/or uncertainty. Even though many papers dealing with such problems have been referenced and analyzed in this work, one has to deepen the analysis of the proposed algorithmic methodologies to solve such complex (V)MVOPs.

\medskip

% Parallel and HPC
There is no substantial work dealing with the design and implementation of massively parallel metaheuristics for (V)MVOPs. Nowadays, supercomputers composed of an important number of heterogeneous devices such as multi-core processors and GPUs are becoming more and more popular. Parallel co-evolutionary metaheuristics represent a promising perspective in the development of efficient and effective algorithms to solve large scale and expensive (V)MVOPs.

\bibliography{article}

\begin{thebibliography}{100}
\providecommand{\url}[1]{#1}
\csname url@samestyle\endcsname
\providecommand{\newblock}{\relax}
\providecommand{\bibinfo}[2]{#2}
\providecommand{\BIBentrySTDinterwordspacing}{\spaceskip=0pt\relax}
\providecommand{\BIBentryALTinterwordstretchfactor}{4}
\providecommand{\BIBentryALTinterwordspacing}{\spaceskip=\fontdimen2\font plus
\BIBentryALTinterwordstretchfactor\fontdimen3\font minus
  \fontdimen4\font\relax}
\providecommand{\BIBforeignlanguage}[2]{{%
\expandafter\ifx\csname l@#1\endcsname\relax
\typeout{** WARNING: IEEEtran.bst: No hyphenation pattern has been}%
\typeout{** loaded for the language `#1'. Using the pattern for}%
\typeout{** the default language instead.}%
\else
\language=\csname l@#1\endcsname
\fi
#2}}
\providecommand{\BIBdecl}{\relax}
\BIBdecl

\bibitem{talbi2021automated}
E.-G. Talbi, ``Automated design of deep neural networks: A survey and unified
  taxonomy,'' \emph{ACM Computing Surveys}, vol.~54, no.~2, 2021.

\bibitem{meng2021comparative}
Z.~Meng, G.~Li, X.~Wang, S.~Sait, and A.~Y{\i}ld{\i}z, ``A comparative study of
  metaheuristic algorithms for reliability-based design optimization
  problems,'' \emph{Archives of Computational Methods in Engineering}, vol.~28,
  no.~3, 2021.

\bibitem{lodi2010mixed}
A.~Lodi, ``Mixed integer programming computation,'' in \emph{50 years of
  integer programming 1958-2008}.\hskip 1em plus 0.5em minus 0.4em\relax
  Springer, 2010.

\bibitem{cooper1981survey}
M.~Cooper, ``A survey of methods for pure nonlinear integer programming,''
  \emph{Management Science}, vol.~27, no.~3, 1981.

\bibitem{pelamatti2021mixed}
J.~Pelamatti, L.~Brevault, M.~Balesdent, E.-G. Talbi, and Y.~Guerin, ``Mixed
  variable gaussian process-based surrogate modeling techniques,''
  \emph{Journal of Aerospace Information Systems}, vol.~18, no.~11, 2021.

\bibitem{martins2021engineering}
J.~Martins and A.~Ning, \emph{Engineering design optimization}.\hskip 1em plus
  0.5em minus 0.4em\relax Cambridge University Press, 2021.

\bibitem{lucidi2005algorithm}
S.~Lucidi, V.~Piccialli, and M.~Sciandrone, ``An algorithm model for mixed
  variable programming,'' \emph{{SIAM} Journal on Optimization}, vol.~15,
  no.~4, 2005.

\bibitem{nyew2015variable}
H.~M. Nyew, O.~Abdelkhalik, and N.~Onder, ``Structured-chromosome evolutionary
  algorithms for variable-size autonomous interplanetary trajectory planning
  optimization,'' \emph{J. Aerosp. Inf. Syst.}, vol.~12, no.~3, 2015.

\bibitem{abdelkhalik2021algorithms}
O.~Abdelkhalik, \emph{Algorithms for Variable-Size Optimization: Applications
  in Space Systems and Renewable Energy}.\hskip 1em plus 0.5em minus
  0.4em\relax CRC Press, 2021.

\bibitem{lampinen1999mixed}
J.~Lampinen and I.~Zelinka, ``Mixed integer-discrete-continuous optimization by
  differential evolution,'' in \emph{5th Int. Conf. on Soft Computing}, 1999.

\bibitem{ho2015improved}
V.~Ho-Huu, T.~Nguyen-Thoi, M.~Nguyen-Thoi, and L.~Le-Anh, ``An improved
  constrained differential evolution using discrete variables for layout
  optimization of truss structures,'' \emph{Expert Systems with Applications},
  vol.~42, no.~20, 2015.

\bibitem{zhao2020ensemble}
F.~Zhao, L.~Zhao, L.~Wang, and H.~Song, ``An ensemble discrete differential
  evolution for the distributed blocking flowshop scheduling with minimizing
  makespan criterion,'' \emph{Expert Systems with Applications}, vol. 160,
  2020.

\bibitem{emmerich2000fixed}
M.~Emmerich, M.~Gr{\"o}tzner, B.~Gro{\ss}, and M.~Sch{\"u}tz, ``Mixed-integer
  evolution strategy for chemical plant optimization with simulators,'' in
  \emph{Evolutionary Design and Manufacture}.\hskip 1em plus 0.5em minus
  0.4em\relax Springer, 2000.

\bibitem{back1991survey}
T.~B{\"a}ck, F.~Hoffmeister, and H.-P. Schwefel, ``A survey of evolution
  strategies,'' in \emph{{ICGA} Fourth Int. Conf. on Genetic Algorithms}, 1991.

\bibitem{miyagi2018well}
A.~Miyagi, Y.~Akimoto, and H.~Yamamoto, ``Well placement optimization for
  carbon dioxide capture and storage via {CMA-ES} with mixed integer support,''
  in \emph{Genetic and Evolutionary Computation Conference Companion}, 2018.

\bibitem{dos2010gaussian}
L.~dos Santos~Coelho, ``Gaussian quantum-behaved {PSO} approaches for
  constrained engineering design problems,'' \emph{Expert Systems with
  Applications}, vol.~37, no.~2, 2010.

\bibitem{kim2020constrained}
T.-H. Kim, M.~Cho, and S.~Shin, ``Constrained mixed-variable design
  optimization based on {PSO} with a diversity classifier for cyclically
  neighboring subpopulations,'' \emph{Mathematics}, vol.~8, no.~11, 2020.

\bibitem{socha2004aco}
K.~Socha, ``Aco for continuous and mixed-variable optimization,'' in \emph{Int.
  Workshop on Ant Colony Optimization and Swarm Intelligence}.\hskip 1em plus
  0.5em minus 0.4em\relax Springer, 2004.

\bibitem{rivas2017coordination}
A.~Rivas and L.~Pareja, ``Coordination of directional overcurrent relays that
  uses an ant colony optimization algorithm for mixed-variable optimization
  problems,'' in \emph{{IEEE} Int. Conf. on Environment and Electrical
  Engineering}.

\bibitem{liao2014fixed}
T.~Liao, K.~Socha, M.~A.~M. de~Oca, T.~St{\"{u}}tzle, and M.~Dorigo, ``{ACO}
  for mixed-variable optimization problems,'' \emph{{IEEE} Trans. Evolutionary
  Computation}, vol.~18, no.~4, 2014.

\bibitem{wang2019estimation}
F.~Wang, Y.~Li, A.~Zhou, and K.~Tang, ``An estimation of distribution algorithm
  for mixed-variable newsvendor problems,'' \emph{{IEEE} Transactions on
  Evolutionary Computation}, vol.~24, no.~3, 2019.

\bibitem{zhou2015estimation}
A.~Zhou, J.~Sun, and Q.~Zhang, ``An estimation of distribution algorithm with
  cheap and expensive local search methods,'' \emph{{IEEE} Transactions on
  Evolutionary Computation}, vol.~19, no.~6, 2015.

\bibitem{jalota2018genetic}
H.~Jalota and M.~Thakur, ``{GA} designed for solving linear or nonlinear
  mixed-integer constrained optimization problems,'' in \emph{Int. Conf. on
  advances in Soft Computing, Intelligent Systems and Applications}, 2018.

\bibitem{liu2018efficient}
D.~Liu, C.~Liu, C.~Zhang, C.~Xu, Z.~Du, and Z.~Wan, ``Efficient hybrid
  algorithms to solve mixed discrete-continuous optimization problems: A
  comparative study,'' \emph{Engineering Computations}, 2018.

\bibitem{maiti2006application}
A.~Maiti, A.~Bhunia, and M.~Maiti, ``An application of real-coded {GA} for
  mixed integer non-linear programming,'' \emph{Applied Mathematics and
  computation}, vol. 183, no.~2, 2006.

\bibitem{exler2008tabu}
O.~Exler, L.~Antelo, J.~Egea, A.~Alonso, and J.~Banga, ``A tabu search-based
  algorithm for mixed-integer nonlinear problems and its application to
  integrated process and control system design,'' \emph{Computers \& Chemical
  Engineering}, vol.~32, no.~8, 2008.

\bibitem{MashinchiOP11}
H.~Mashinchi, M.~Orgun, and W.~Pedrycz, ``Hybrid optimization with improved
  tabu search,'' \emph{Applied Soft Computing}, vol.~11, no.~2, 2011.

\bibitem{zhang1993mixed}
C.~Zhang and H.-P. Wang, ``Mixed-discrete nonlinear optimization with simulated
  annealing,'' \emph{Engineering Optimization}, vol.~21, no.~4, 1993.

\bibitem{koken2018simulated}
P.~Koken, H.~Seok, and S.~Yoon, ``A simulated annealing algorithm with
  neighbourhood list for capacitated dynamic lot-sizing problem with returns
  and hybrid products,'' \emph{International Journal of Computer Integrated
  Manufacturing}, vol.~31, no.~8, 2018.

\bibitem{mohan1999controlled}
C.~Mohan and H.~Nguyen, ``A controlled random search technique incorporating
  the simulated annealing concept for solving integer and mixed integer global
  optimization problems,'' \emph{Computational Optimization and Applications},
  vol.~14, no.~1, 1999.

\bibitem{akay2021survey}
B.~Akay, D.~Karaboga, B.~Gorkemli, and E.~Kaya, ``A survey on the artificial
  bee colony algorithm variants for binary, integer and mixed integer
  programming problems,'' \emph{Applied Soft Computing}, vol. 106, 2021.

\bibitem{audet2001pattern}
C.~Audet and J.~E. Dennis, ``Pattern search algorithms for mixed variable
  programming,'' \emph{SIAM Journal on Optimization}, vol.~11, no.~3, 2001.

\bibitem{cheung1997coupling}
B.~Cheung, K.~Langevin, and H.~Delmaire, ``Coupling genetic algorithm with a
  grid search method to solve mixed integer nonlinear programming problems,''
  \emph{Computers \& Mathematics with Applications}, vol.~34, no.~12, 1997.

\bibitem{pelamatti2017deal}
J.~Pelamatti, L.~Brevault, M.~Balesdent, E.-G. Talbi, and Y.~Guerin, ``How to
  deal with mixed-variable optimization problems,'' in \emph{World Congress of
  Structural and Multidisciplinary Optimisation}, 2017.

\bibitem{Talbi2009}
E.-G. Talbi, \emph{Metaheuristics: from design to implementation}.\hskip 1em
  plus 0.5em minus 0.4em\relax Wiley, 2009.

\bibitem{del2019bio}
v.~y. n.~p. J.~Del Ser~et al., journal={Swarm and Evolutionary Computation},
  ``Bio-inspired computation: Where we stand and what's next.''

\bibitem{storn1997DE}
R.~Storn and K.~V. Price, ``Differential evolution: A simple and efficient
  heuristic for global optimization over continuous spaces,'' \emph{Journal of
  Global Optimization}, vol.~11, no.~4, 1997.

\bibitem{eberhart2001swarm}
R.~Eberhart, Y.~Shi, and J.~Kennedy, \emph{Swarm intelligence}.\hskip 1em plus
  0.5em minus 0.4em\relax Elsevier, 2001.

\bibitem{stork2020new}
J.~Stork, A.~E. Eiben, and T.~Bartz-Beielstein, ``A new taxonomy of global
  optimization algorithms,'' \emph{Natural Computing}, 2020.

\bibitem{yildiz2020comparative}
A.~Yildiz, H.~Abderazek, and S.~Mirjalili, ``A comparative study of recent
  non-traditional methods for mechanical design optimization,'' \emph{Archives
  of Computational Methods in Engineering}, vol.~27, no.~4, 2020.

\bibitem{gupta2021comparison}
S.~Gupta, H.~Abderazek, B.~Y{\i}ld{\i}z, A.~Yildiz, S.~Mirjalili, and S.~Sait,
  ``Comparison of metaheuristic optimization algorithms for solving constrained
  mechanical design optimization problems,'' \emph{Expert Systems with
  Applications}, vol. 183, 2021.

\bibitem{kania2016solving}
A.~Kania and K.~Sidarto, ``Solving mixed integer nonlinear programming problems
  using spiral dynamics optimization algorithm,'' in \emph{AIP Conference},
  vol. 1716, no.~1, 2016.

\bibitem{angira2006optimization}
R.~Angira and B.~Babu, ``Optimization of process synthesis and design problems:
  A modified differential evolution approach,'' \emph{Chemical Engineering
  Science}, vol.~61, no.~14, 2006.

\bibitem{ali2020novel}
I.~Ali, D.~Essam, and K.~Kasmarik, ``A novel design of differential evolution
  for solving discrete traveling salesman problems,'' \emph{Swarm and
  Evolutionary Computation}, vol.~52, 2020.

\bibitem{zheng2019differential}
S.~Zheng, B.~Xiang, X.~Zhang, and J.~Zhang, ``Differential evolution
  optimization algorithm for antenna designs with mixed discrete-continuous
  variables,'' in \emph{Int. Conf. on Microwave and Millimeter Wave Technology
  (ICMMT)}, 2019.

\bibitem{mohamed2017efficient}
A.~W. Mohamed, ``An efficient modified differential evolution algorithm for
  solving constrained non-linear integer and mixed-integer global optimization
  problems,'' \emph{International Journal of Machine Learning and Cybernetics},
  vol.~8, no.~3, 2017.

\bibitem{ponsich2011differential}
A.~Ponsich and C.~Coello, ``Differential evolution performances for the
  solution of mixed-integer constrained process engineering problems,''
  \emph{Applied Soft Computing}, vol.~11, no.~1, 2011.

\bibitem{varadarajan2008differential}
M.~Varadarajan and K.~Swarup, ``Differential evolution approach for optimal
  reactive power dispatch,'' \emph{Applied soft computing}, vol.~8, no.~4,
  2008.

\bibitem{mohamed2019solving}
A.~Mohamed, A.~Mohamed, E.~Elfeky, and M.~Saleh, ``Solving constrained
  non-linear integer and mixed-integer global optimization problems using
  enhanced directed differential evolution algorithm,'' in \emph{Machine
  Learning Paradigms: Theory and Application}.\hskip 1em plus 0.5em minus
  0.4em\relax Springer, 2019.

\bibitem{lin2004mixed}
Y.-C. Lin, K.-S. Hwang, and F.-S. Wang, ``A mixed-coding scheme of evolutionary
  algorithms to solve mixed-integer nonlinear programming problems,''
  \emph{Computers \& Mathematics with Applications}, vol.~47, no.~8, 2004.

\bibitem{peng2021multi}
H.~Peng, Y.~Han, C.~Deng, J.~Wang, and Z.~Wu, ``Multi-strategy co-evolutionary
  differential evolution for mixed-variable optimization,''
  \emph{Knowledge-Based Systems}, vol. 229, 2021.

\bibitem{rao2005hybrid}
S.~Rao and Y.~Xiong, ``A hybrid {GA} for mixed-discrete design optimization,''
  \emph{Journal of Mechanical Design}, vol. 127, no.~6, 2005.

\bibitem{deep2009real}
K.~Deep, K.~Singh, M.~Kansal, and C.~Mohan, ``A real coded genetic algorithm
  for solving integer and mixed integer optimization problems,'' \emph{Applied
  Mathematics and Computation}, vol. 212, no.~2, 2009.

\bibitem{wu1995genetic}
S.-J. Wu and P.-T. Chow, ``{GAs} for nonlinear mixed discrete-integer
  optimization problems via meta-genetic parameter optimization,''
  \emph{Engineering Optimization}, vol.~24, no.~2, 1995.

\bibitem{yokota1996genetic}
T.~Yokota, M.~Gen, and Y.-X. Li, ``Genetic algorithm for non-linear mixed
  integer programming problems and its applications,'' \emph{Computers \&
  Industrial Engineering}, vol.~30, no.~4, 1996.

\bibitem{yan2004solving}
L.~Yan, K.~Shen, and S.~Hu, ``Solving mixed integer nonlinear programming
  problems with line-up competition algorithm,'' \emph{Computers \& Chemical
  Engineering}, vol.~28, no.~12, 2004.

\bibitem{jun2013improved}
W.~Jun, G.~Yuelin, and Y.~Lina, ``An improved differential evolution algorithm
  for mixed integer programming problems,'' in \emph{{IEEE} Int. Conf. on
  Computational Intelligence and Security}, 2013.

\bibitem{lin2001co}
Y.-C. Lin, K.-S. Hwang, and F.-S. Wang, ``Co-evolutionary hybrid differential
  evolution for mixed-integer optimization problems,'' \emph{Engineering
  Optimization}, vol.~33, no.~6, 2001.

\bibitem{guo2004swarm}
C.-X. Guo, J.-S. Hu, B.~Ye, and Y.-J. Cao, ``Swarm intelligence for
  mixed-variable design optimization,'' \emph{Journal of Zhejiang
  University-SCIENCE A}, vol.~5, no.~7, 2004.

\bibitem{nahvi2011particle}
H.~Nahvi and I.~Mohagheghian, ``A particle swarm optimization algorithm for
  mixed variable nonlinear problems,'' 2011.

\bibitem{chowdhury2013mixed}
S.~Chowdhury, W.~Tong, A.~Messac, and J.~Zhang, ``A mixed-discrete particle
  swarm optimization algorithm with explicit diversity-preservation,''
  \emph{Structural and Multidisciplinary Optimization}, vol.~47, no.~3, 2013.

\bibitem{venter2004multidisciplinary}
G.~Venter and J.~Sobieszczanski-Sobieski, ``Multidisciplinary optimization of a
  transport aircraft wing using {PSO},'' \emph{Structural and Multidisciplinary
  optimization}, vol.~26, no. 1-2, 2004.

\bibitem{parsopoulos2002recent}
K.~Parsopoulos and M.~Vrahatis, ``Recent approaches to global optimization
  problems through {PSO},'' \emph{Natural Computing}, vol.~1, no.~2, 2002.

\bibitem{kanagaraj2014effective}
G.~Kanagaraj, S.~Ponnambalam, N.~Jawahar, and M.~Nilakantan, ``An effective
  hybrid cuckoo search and genetic algorithm for constrained engineering design
  optimization,'' \emph{Engineering Optimization}, vol.~46, no.~10, 2014.

\bibitem{gandomi2013cuckoo}
A.~Gandomi, X.-S. Yang, and A.~Alavi, ``Cuckoo search algorithm: a
  metaheuristic approach to solve structural optimization problems,''
  \emph{Engineering with Computers}, vol.~29, no.~1, 2013.

\bibitem{mahdavi2007improved}
M.~Mahdavi, M.~Fesanghary, and E.~Damangir, ``An improved harmony search
  algorithm for solving optimization problems,'' \emph{Applied mathematics and
  computation}, vol. 188, no.~2, 2007.

\bibitem{gupta2019efficient}
S.~Gupta and K.~Deep, ``An efficient grey wolf optimizer with opposition-based
  learning and chaotic local search for integer and mixed-integer optimization
  problems,'' \emph{Journal for Science \& Engineering}, vol.~44, no.~8, 2019.

\bibitem{rao2011teaching}
V.~Rao, V.~Savsani, and D.~Vakharia, ``Teaching--learning-based optimization: a
  novel method for constrained mechanical design optimization problems,''
  \emph{Computer-aided design}, vol.~43, no.~3, 2011.

\bibitem{talatahari2020discrete}
S.~Talatahari and V.~Goodarzimehr, ``A discrete hybrid teaching-learning-based
  optimization algorithm for optimization of space trusses,'' \emph{Journal of
  Structural Engineering and Geo-Techniques}, vol.~10, no.~1, 2020.

\bibitem{das2017transmission}
S.~Das, A.~Verma, and P.~Bijwe, ``Transmission network expansion planning using
  a modified artificial bee colony algorithm,'' \emph{International
  Transactions on Electrical Energy Systems}, vol.~27, no.~9, 2017.

\bibitem{gandomi2011mixed}
A.~Gandomi, X.-S. Yang, and A.~Alavi, ``Mixed variable structural optimization
  using firefly algorithm,'' \emph{Computers \& Structures}, vol.~89, no.
  23-24, 2011.

\bibitem{SchluterEB09}
M.~Schl{\"{u}}ter, J.~Egea, and J.~Banga, ``Extended ant colony optimization
  for non-convex mixed integer nonlinear programming,'' \emph{Comput. Oper.
  Res.}, vol.~36, no.~7, 2009.

\bibitem{LiaoKH12}
T.~Liao, R.~Kuo, and J.~Hu, ``Hybrid {ACO} algorithms for mixed
  discrete-continuous optimization problems,'' \emph{Applied Mathematics and
  Computation}, vol. 219, no.~6, 2012.

\bibitem{miao2016modified}
Y.~Miao, Q.~Su, Z.~Hu, and X.~Xia, ``Modified differential evolution algorithm
  with onlooker bee operator for mixed discrete-continuous optimization,''
  \emph{SpringerPlus}, vol.~5, no.~1, 2016.

\bibitem{liao2010two}
W.~Liao, ``Two hybrid differential evolution algorithms for engineering design
  optimization,'' \emph{Applied Soft Computing}, vol.~10, no.~4, 2010.

\bibitem{dos2008use}
L.~C. dos Santos and V.~Mariani, ``Use of chaotic sequences in a biologically
  inspired algorithm for engineering design optimization,'' \emph{Expert
  Systems with Applications}, vol.~34, no.~3, 2008.

\bibitem{dos2011chaotic}
C.~L. dos Santos, B.~de~Andrade, and V.~Mariani, ``A chaotic firefly algorithm
  applied to reliability-redundancy optimization,'' in \emph{{IEEE} Congress of
  Evolutionary Computation (CEC)}, 2011.

\bibitem{srinivas2007differential}
M.~Srinivas and G.~Rangaiah, ``Differential evolution with tabu list for
  solving nonlinear and mixed-integer nonlinear programming problems,''
  \emph{Industrial \& engineering chemistry research}, vol.~46, no.~22, 2007.

\bibitem{yi2013three}
H.~Yi, Q.~Duan, and W.~Liao, ``Three improved hybrid metaheuristic algorithms
  for engineering design optimization,'' \emph{Applied Soft Computing},
  vol.~13, no.~5, 2013.

\bibitem{garg2016hybrid}
H.~Garg, ``A hybrid {PSO-GA} algorithm for constrained optimization problems,''
  \emph{Applied Mathematics and Computation}, vol. 274, 2016.

\bibitem{rao2019engineering}
S.~Rao, \emph{Engineering optimization: theory and practice}.\hskip 1em plus
  0.5em minus 0.4em\relax John Wiley \& Sons, 2019.

\bibitem{kitayama2006method}
S.~Kitayama and K.~Yasuda, ``A method for mixed integer programming problems by
  {PSO},'' \emph{Electrical Engineering in Japan}, vol. 157, no.~2, 2006.

\bibitem{lotfipour2016discrete}
A.~Lotfipour and H.~Afrakhte, ``A discrete teaching-learning-based optimization
  algorithm to solve distribution system reconfiguration in presence of
  distributed generation,'' \emph{International Journal of Electrical Power \&
  Energy systems}, vol.~82, 2016.

\bibitem{chowdhury2010developing}
S.~Chowdhury, A.~Messac, and R.~Khire, ``Developing a non-gradient based
  mixed-discrete optimization approach for comprehensive product platform
  planning (cp3),'' in \emph{13th AIAA/ISSMO Multidisciplinary Analysis
  Optimization Conference}, 2010.

\bibitem{fu1991mixed}
J.-F. Fu, R.~Fenton, and W.~Cleghorn, ``A mixed integer-discrete-continuous
  programming method and its application to engineering design optimization,''
  \emph{Engineering Optimization}, vol.~17, no.~4, 1991.

\bibitem{murray2010algorithm}
W.~Murray and K.-M. Ng, ``An algorithm for nonlinear optimization problems with
  binary variables,'' \emph{Computational optimization and applications},
  vol.~47, no.~2, 2010.

\bibitem{abhishek2010modeling}
K.~Abhishek, S.~Leyffer, and J.~Linderoth, ``Modeling without categorical
  variables,'' \emph{Optimization and Engineering}, vol.~11, no.~2, 2010.

\bibitem{he2004improved}
S.~He, E.~Prempain, and Q.~Wu, ``An improved {PSO} for mechanical design
  optimization problems,'' \emph{Engineering optimization}, vol.~36, no.~5,
  2004.

\bibitem{schmidt2005combined}
H.~Schmidt and G.~Thierauf, ``A combined heuristic optimization technique,''
  \emph{Advances in Engineering Software}, vol.~36, no.~1, 2005.

\bibitem{crawford2017putting}
B.~Crawford, R.~Soto, G.~Astorga, J.~Garc{\'\i}a, C.~Castro, and F.~Paredes,
  ``Putting continuous metaheuristics to work in binary search spaces,''
  \emph{Complexity}, vol. 2017, 2017.

\bibitem{yu2016stock}
L.~Yu, L.~Hu, and L.~Tang, ``Stock selection with a novel sigmoid-based mixed
  discrete-continuous differential evolution algorithm,'' \emph{IEEE
  Transactions on Knowledge and Data Engineering}, vol.~28, no.~7, 2016.

\bibitem{kennedy1997discrete}
J.~Kennedy and R.~Eberhart, ``A discrete binary version of the particle swarm
  algorithm,'' in \emph{{IEEE} Int. Conf. on Systems, Man, and Cybernetics},
  vol.~5, 1997.

\bibitem{palit2011cryptanalytic}
S.~Palit, S.~Sinha, M.~Molla, A.~Khanra, and M.~Kule, ``A cryptanalytic attack
  on the knapsack cryptosystem using binary firefly algorithm,'' in
  \emph{{IEEE} Int. Conf. on Computer and Communication Technology (ICCCT)},
  2011.

\bibitem{pampara2006binary}
G.~Pampara, A.~Engelbrecht, and N.~Franken, ``Binary differential evolution,''
  in \emph{{IEEE} Int. Conf. on Evolutionary Computation}, 2006.

\bibitem{LiuWJ07}
B.~Liu, L.~Wang, and Y.~Jin, ``An effective pso-based memetic algorithm for
  flow shop scheduling,'' \emph{{IEEE} Trans. Syst. Man Cybern. Part {B}},
  vol.~37, no.~1, 2007.

\bibitem{ali2016differential}
I.~Ali, S.~Elsayed, T.~Ray, and R.~Sarker, ``A differential evolution algorithm
  for solving resource constrained project scheduling problems,'' in
  \emph{{ACALCI'2016} Conference on Artificial Life and Computational
  Intelligence}, 2016.

\bibitem{ouaarab2015random}
A.~Ouaarab, B.~Ahiod, and X.-S. Yang, ``Random-key cuckoo search for the
  travelling salesman problem,'' \emph{Soft Computing}, vol.~19, no.~4, 2015.

\bibitem{hafiz2016particle}
F.~Hafiz and A.~Abdennour, ``{PSO} algorithm variants for the quadratic
  assignment problems - a probabilistic learning approach,'' \emph{Expert
  Systems with Applications}, vol.~44, 2016.

\bibitem{tasgetiren2007particle}
F.~Tasgetiren, Y.-C. Liang, M.~Sevkli, and G.~Gencyilmaz, ``A {PSO} algorithm
  for makespan and total flowtime minimization in the permutation flowshop
  sequencing problem,'' \emph{European Journal of Operational Research}, vol.
  177, no.~3, 2007.

\bibitem{ulker2016adaptation}
E.~{\"U}lker, ``Adaptation of harmony search algorithm for dna fragment
  assembly problem,'' in \emph{{IEEE} Computing Conference (SAI)}, 2016.

\bibitem{yousif2014discrete}
A.~Yousif, S.~Nor, A.~Abdullah, and M.~Bashir, ``A discrete firefly algorithm
  for scheduling jobs on computational grid,'' in \emph{Cuckoo Search and
  Firefly Algorithm}.

\bibitem{kumar2011design}
A.~Kumar and S.~Chakarverty, ``Design optimization for reliable embedded system
  using cuckoo search,'' in \emph{{IEEE} Int. Conf. on Electronics Computer
  Technology}, vol.~1, 2011.

\bibitem{li2013hybrid}
X.~Li and M.~Yin, ``A hybrid cuckoo search via l{\'e}vy flights for the
  permutation flow shop scheduling problem,'' \emph{International Journal of
  Production Research}, vol.~51, no.~16, 2013.

\bibitem{arora1994methods}
J.~Arora, M.~Huang, and C.~Hsieh, ``Methods for optimization of nonlinear
  problems with discrete variables: a review,'' \emph{Structural optimization},
  vol.~8, no.~2, 1994.

\bibitem{davydov1972application}
E.~Davydov and I.~Sigal, ``Application of penalty function method in integer
  programming problems,'' \emph{Engineering Cybernetics}, vol.~10, no.~1, 1972.

\bibitem{shin1990penalty}
D.~Shin, Z.~G{\"u}rdal, and J.~Griffin, ``A penalty approach for nonlinear
  optimization with discrete design variables,'' \emph{Engineering
  Optimization}, vol.~16, no.~1, 1990.

\bibitem{Li92}
H.-L. Li, ``An approximate method for local optima for nonlinear mixed integer
  programming problems,'' \emph{Comput. Oper. Res.}, vol.~19, no.~5, 1992.

\bibitem{kitayama2006penalty}
S.~Kitayama, M.~Arakawa, and K.~Yamazaki, ``Penalty function approach for the
  mixed discrete nonlinear problems by {PSO},'' \emph{Structural and
  Multidisciplinary Optimization}, vol.~32, no.~3, 2006.

\bibitem{GoldbergDC92}
D.~Goldberg, K.~Deb, and J.~H. Clark, ``Genetic algorithms, noise, and the
  sizing of populations,'' \emph{Complex Systems}, vol.~6, no.~4, 1992.

\bibitem{michalewicz1996genetic}
Z.~Michalewicz and Z.~Michalewicz, \emph{Genetic algorithms+ data structures=
  evolution programs}.\hskip 1em plus 0.5em minus 0.4em\relax Springer Science
  \& Business Media, 1996.

\bibitem{dimopoulos2007mixed}
G.~Dimopoulos, ``Mixed-variable engineering optimization based on evolutionary
  and social metaphors,'' \emph{Computer methods in applied mechanics and
  engineering}, vol. 196, no. 4-6, 2007.

\bibitem{costa2001evolutionary}
L.~Costa and P.~Oliveira, ``Evolutionary algorithms approach to the solution of
  mixed integer non-linear programming problems,'' \emph{Computers \& Chemical
  Engineering}, vol.~25, no. 2-3, 2001.

\bibitem{turkkan2003discrete}
N.~Turkkan, ``Discrete optimization of structures using a floating-point
  genetic algorithm,'' in \emph{Annual Conf. of the Canadian Society for Civil
  Engineering}, 2003.

\bibitem{lin1992genetic}
C.-Y. Lin and P.~Hajela, ``{GAs} in optimization problems with discrete and
  integer design variables,'' \emph{Engineering optimization}, vol.~19, no.~4,
  1992.

\bibitem{stelmack1998genetic}
M.~Stelmack, N.~Nakashima, and S.~Batill, ``Genetic algorithms for mixed
  discrete/continuous optimization in multidisciplinary design,'' in \emph{7th
  AIAA/USAF/NASA/ISSMO Symposium on Multidisciplinary Analysis and
  Optimization}, 1998.

\bibitem{Coelho10}
L.~Coelho, ``Gaussian quantum-behaved particle swarm optimization approaches
  for constrained engineering design problems,'' \emph{Expert Systems with
  Applications}, vol.~37, no.~2, 2010.

\bibitem{wang2008ranking}
J.~Wang and Z.~Yin, ``A ranking selection-based particle swarm optimizer for
  engineering design optimization problems,'' \emph{Structural and
  multidisciplinary optimization}, vol.~37, no.~2, 2008.

\bibitem{de2019hybridizing}
A.~De, J.~Wang, and M.~Tiwari, ``Hybridizing basic variable neighborhood search
  with {PSO} for solving sustainable ship routing and bunker management
  problem,'' \emph{IEEE Transactions on Intelligent Transportation Systems},
  vol.~21, no.~3, 2019.

\bibitem{gardi2011local}
F.~Gardi and K.~Nouioua, ``Local search for mixed-integer nonlinear
  optimization: a methodology and an application,'' in \emph{European
  Conference on Evolutionary Computation in Combinatorial Optimization}, 2011.

\bibitem{CardosaSA}
M.~F. Cardoso, R.~L. Salcedo, F.~de~Azevedo, and D.~Barbosa, ``A simulated
  annealing approach to the solution of {MINLP} problems,'' \emph{Computers \&
  Chemical Engineering}, vol.~21, no.~12, 1997.

\bibitem{olsson1975nelder}
D.~Olsson and L.~Nelson, ``The nelder-mead simplex procedure for function
  minimization,'' \emph{Technometrics}, vol.~17, no.~1, 1975.

\bibitem{rudolph1994evolutionary}
R.~G{\"u}nter, ``An evolutionary algorithm for integer programming,'' in
  \emph{Int. Conf. on Parallel Problem Solving from Nature}.\hskip 1em plus
  0.5em minus 0.4em\relax Springer, 1994.

\bibitem{van2016multicriteria}
K.~V. der Blom, S.~Boonstra, H.~Hofmeyer, and M.~Emmerich, ``Multicriteria
  building spatial design with mixed integer evolutionary algorithms,'' in
  \emph{Int. Conf. on Parallel Problem Solving from Nature}.\hskip 1em plus
  0.5em minus 0.4em\relax Springer, 2016.

\bibitem{li2014discrete}
H.~Li and L.~Zhang, ``A discrete hybrid differential evolution algorithm for
  solving integer programming problems,'' \emph{Engineering Optimization},
  vol.~46, no.~9, 2014.

\bibitem{li2013mixed}
R.~L. et~al., ``Mixed integer evolution strategies for parameter
  optimization,'' \emph{Evolutionary Computation}, vol.~21, no.~1, 2013.

\bibitem{lin2018hybrid}
Y.~Lin, Y.~Liu, W.-N. Chen, and J.~Zhang, ``A hybrid differential evolution
  algorithm for mixed-variable optimization problems,'' \emph{Information
  Sciences}, vol. 466, 2018.

\bibitem{datta2013real}
D.~Datta and J.~Figueira, ``A real integer discrete-coded differential
  evolution,'' \emph{Applied Soft Computing}, vol.~13, no.~9, 2013.

\bibitem{li2008fixed}
R.~L. et~al., ``Metamodel-assisted mixed integer evolution strategies and their
  application to intravascular ultrasound image analysis,'' in \emph{{IEEE}
  Congress on Evolutionary Computation {CEC}}, 2008.

\bibitem{beyer2002evolution}
H.-G. Beyer and H.-P. Schwefel, ``Evolution strategies - a comprehensive
  introduction,'' \emph{Natural computing}, vol.~1, no.~1, 2002.

\bibitem{back1996evolutionary}
T.~Back, \emph{Evolutionary algorithms in theory and practice: evolution
  strategies, evolutionary programming, genetic algorithms}.\hskip 1em plus
  0.5em minus 0.4em\relax Oxford university press, 1996.

\bibitem{CAO2000931}
Y.~J. Cao, L.~Jiang, and Q.~H. Wu, ``An evolutionary programming approach to
  mixed-variable optimization problems,'' \emph{Applied Mathematical
  Modelling}, vol.~24, no.~12, 2000.

\bibitem{kincaid2004bell}
R.~Kincaid, M.~Griffith, R.~Sykes, and J.~Sobieszczanski-Sobieski, ``Bell-curve
  {GA} for mixed continuous and discrete optimization problems,''
  \emph{Structural and Multidisciplinary Optimization}, vol.~26, no.~6, 2004.

\bibitem{gao2010comprehensive}
L.~Gao and A.~Hailu, ``Comprehensive learning particle swarm optimizer for
  constrained mixed-variable optimization problems,'' \emph{International
  Journal of Computational Intelligence Systems}, vol.~3, no.~6, 2010.

\bibitem{wang2021particle}
F.~Wang, H.~Zhang, and A.~Zhou, ``A particle swarm optimization algorithm for
  mixed-variable optimization problems,'' \emph{Swarm and Evolutionary
  Computation}, vol.~60, 2021.

\bibitem{mokarram2018new}
V.~Mokarram and M.~Banan, ``A new pso-based algorithm for multi-objective
  optimization with continuous and discrete design variables,''
  \emph{Structural and Multidisciplinary Optimization}, vol.~57, no.~2, 2018.

\bibitem{hinojosa2013modeling}
V.~Hinojosa and R.~Araya, ``Modeling a mixed-integer-binary small-population
  evolutionary {PSO} for solving the optimal power flow problem in electric
  power systems,'' \emph{Applied Soft Computing}, vol.~13, no.~9, 2013.

\bibitem{sun2011modified}
C.~Sun, J.~Zeng, and J.-S. Pan, ``A modified {PSO} with feasibility-based rules
  for mixed-variable optimization problems,'' \emph{International Journal of
  Innovative Computing, Information and Control}, vol.~7, no.~6, 2011.

\bibitem{rezaee2020mixed}
J.~Rezaee, ``A mixed binary-continuous {PSO} algorithm for unit commitment in
  microgrids considering uncertainties and emissions,'' \emph{International
  Transactions on Electrical Energy Systems}, vol.~30, no.~11, 2020.

\bibitem{liao2007fixed}
C.~Liao, C.~Tseng, and P.~Luarn, ``A discrete version of particle swarm
  optimization for flowshop scheduling problems,'' \emph{Comput. Oper. Res.},
  vol.~34, no.~10, 2007.

\bibitem{yiqing2007improved}
L.~Yiqing, L.~Xigang, and L.~Yongjian, ``An improved {PSO} algorithm for
  solving non-convex {NLP/MINLP} problems with equality constraints,''
  \emph{Computers \& Chemical Engineering}, vol.~31, no.~3, 2007.

\bibitem{wang2017hybrid}
L.~Wang, J.~Pei, M.~Menhas, J.~Pi, M.~Fei, and P.~Pardalos, ``A hybrid-coded
  human learning optimization for mixed-variable optimization problems,''
  \emph{Knowledge-Based Systems}, vol. 127, 2017.

\bibitem{larranaga1999optimization}
P.~Larranag, R.~Etxeberria, J.~Lozano, and J.~Pena, ``Optimization by learning
  and simulation of bayesian and gaussian networks. university of the basque
  country technical report ehu-kzaa-ik-4/99,'' 1999.

\bibitem{bosman2000mixed}
P.~Bosman and D.~Thierens, ``Mixed ideas,'' \emph{Utrecht UniversityTechnical
  Report UU-CS-2000-45. Utrecht University, Utrecht, Netherlands}, 2000.

\bibitem{ocenasek2002estimation}
J.~Ocenasek and J.~Schwarz, ``Estimation of distribution algorithm for mixed
  continuous-discrete optimization problems,'' in \emph{2nd Euro-Int. Symp. on
  Computational Intelligence}.\hskip 1em plus 0.5em minus 0.4em\relax IOS Press
  Kosice, 2002.

\bibitem{sahoo2014efficient}
L.~Sahoo, A.~Banerjee, A.~Bhunia, and S.~Chattopadhyay, ``An efficient {GA-PSO}
  approach for solving mixed-integer nonlinear programming problem in
  reliability optimization,'' \emph{Swarm and Evolutionary Computation},
  vol.~19, 2014.

\bibitem{gao2016difference}
Y.~Gao, Y.~Sun, and J.~Wu, ``Difference-genetic co-evolutionary algorithm for
  nonlinear mixed integer programming problems,'' \emph{Journal of Nonlinear
  Science and Its Applications}, vol.~9, no.~3, 2016.

\bibitem{hedar2011filter}
A.-R. Hedar and A.~Fahim, ``Filter-based {GA} for mixed variable programming,''
  \emph{Numerical Algebra, Control \& Optimization}, vol.~1, no.~1, 2011.

\bibitem{shi2017adaptive}
W.~Shi, W.-N. Chen, Y.~Lin, T.~Gu, S.~Kwong, and J.~Zhang, ``An adaptive
  estimation of distribution algorithm for multipolicy insurance investment
  planning,'' \emph{{IEEE} Transactions on Evolutionary Computation}, vol.~23,
  no.~1, 2017.

\bibitem{li2021sizing}
W.~Li, G.~Zhang, X.~Yang, Z.~Tao, and H.~Xu, ``Sizing a hybrid renewable energy
  system by a coevolutionary multiobjective optimization algorithm,''
  \emph{Complexity}, vol. 2021, 2021.

\bibitem{abramson2009mesh}
M.~A. Abramson, C.~Audet, J.~W. Chrissis, and J.~G. Walston, ``Mesh adaptive
  direct search algorithms for mixed variable optimization,''
  \emph{Optimization Letters}, vol.~3, no.~1, 2009.

\bibitem{abramson2004filter}
M.~A. Abramson, C.~Audet, and J.~E. Dennis, ``Filter pattern search algorithms
  for mixed variable constrained optimization problems,'' Rice University,
  Tech. Rep., 2004.

\bibitem{abramson2003pattern}
M.~A. Abramson, \emph{Pattern search algorithms for mixed variable general
  constrained optimization problems}.\hskip 1em plus 0.5em minus 0.4em\relax
  Rice University, 2003.

\bibitem{audet2006sequential}
C.~Audet and J.~E. Dennis, ``Mesh adaptive direct search algorithms for
  constrained optimization,'' \emph{{SIAM} Journal on Optimization}, vol.~17,
  no.~1, 2006.

\bibitem{capitanescu2010sensitivity}
F.~Capitanescu and L.~Wehenkel, ``Sensitivity-based approaches for handling
  discrete variables in optimal power flow computations,'' \emph{{IEEE}
  Transactions on Power Systems}, vol.~25, no.~4, 2010.

\bibitem{yan2006hybrid}
W.~Yan, F.~Liu, C.~Chung, and K.~Wong, ``A hybrid {GA}-interior point method
  for optimal reactive power flow,'' \emph{IEEE Transactions on Power Systems},
  vol.~21, no.~3, 2006.

\bibitem{gao2014hybrid}
Y.~Gao, X.~Chang, and J.~Liu, ``Hybrid coding collaborative {DE-ACO} algorithm
  for solving mixed-integer programming problems,'' \emph{Journal of
  Computing}, vol.~9, no.~1, 2014.

\bibitem{stelmack1997concurrent}
M.~Stelmack, S.~Batill, M.~Stelmack, and S.~Batill, ``Concurrent subspace
  optimization of mixed continuous/discrete systems,'' in \emph{38th
  Structures, Structural Dynamics, and Materials Conf.}, 1997.

\bibitem{lin2013mixed}
Y.~Lin, ``Mixed-integer constrained optimization based on memetic algorithm,''
  \emph{Journal of Applied Research and Technology}, vol.~11, no.~2, 2013.

\bibitem{nema}
S.~Nema, J.~Goulermas, G.~Sparrow, and P.~Cook, ``A hybrid particle swarm
  branch-and-bound optimizer for mixed discrete nonlinear programming,''
  \emph{IEEE Transactions on Systems, Man, and Cybernetics - Part A: Systems
  and Humans}, vol.~38, no.~6, 2008.

\bibitem{hua2006effective}
Z.~Hua and F.~Huang, ``An effective genetic algorithm approach to large scale
  mixed integer programming problems,'' \emph{Applied Mathematics and
  Computation}, vol. 174, no.~2, 2006.

\bibitem{praharaj1992two}
S.~Praharaj and S.~Azarm, ``Two-level nonlinear mixed discrete-continuous
  optimization-based design,'' \emph{Adv. Design Autom.}, vol.~1, no.~44, 1992.

\bibitem{talbi2013taxonomy}
E.-G. Talbi, ``A taxonomy of metaheuristics for bi-level optimization,'' in
  \emph{Metaheuristics for bi-level optimization}.\hskip 1em plus 0.5em minus
  0.4em\relax Springer, 2013.

\bibitem{talbi2013}
------, \emph{Metaheuristics for bilevel optimization}.\hskip 1em plus 0.5em
  minus 0.4em\relax Springer, 2013.

\bibitem{sinha2017review}
A.~Sinha, P.~Malo, and K.~Deb, ``A review on bilevel optimization: from
  classical to evolutionary approaches and applications,'' \emph{{IEEE}
  Transactions on Evolutionary Computation}, vol.~22, no.~2, 2017.

\bibitem{vanderbeck2006generic}
F.~Vanderbeck and M.~Savelsbergh, ``A generic view of dantzig--wolfe
  decomposition in mixed integer programming,'' \emph{Operations Research
  Letters}, vol.~34, no.~3, 2006.

\bibitem{geoffrion1972generalized}
A.~Geoffrion, ``Generalized benders decomposition,'' \emph{Journal of
  Optimization Theory and Applications}, vol.~10, no.~4, 1972.

\bibitem{duran1986outer}
M.~Duran and I.~Grossmann, ``An outer-approximation algorithm for a class of
  mixed-integer nonlinear programs,'' \emph{Mathematical Programming}, vol.~36,
  no.~3, 1986.

\bibitem{floudas1995nonlinear}
C.~Floudas, \emph{Nonlinear and mixed-integer optimization: fundamentals and
  applications}.\hskip 1em plus 0.5em minus 0.4em\relax Oxford University
  Press, 1995.

\bibitem{chanthasuwannasin2017mixed}
M.~Chanthasuwannasin, B.~Kottititum, and T.~Srinophakun, ``A mixed coding
  scheme of a particle swarm optimization,'' \emph{Chemical Engineering
  Communications}, vol. 204, no.~8, 2017.

\bibitem{roy2019mixed}
S.~Roy, W.~Crossley, B.~Stanford, K.~Moore, and J.~S. Gray, ``A mixed integer
  efficient global optimization algorithm with multiple infill strategy-applied
  to a wing topology optimization problem,'' in \emph{AIAA Scitech 2019 Forum},
  2019.

\bibitem{wang2011simultaneous}
W.~Wang, S.~Guo, and W.~Yang, ``Simultaneous partial topology and size
  optimization of a wing structure using ant colony and gradient based
  methods,'' \emph{Engineering Optimization}, vol.~43, no.~4, 2011.

\bibitem{jones1998efficient}
D.~Jones, M.~Schonlau, and W.~Welch, ``Efficient global optimization of
  expensive black-box functions,'' \emph{Journal of Global Optimization},
  vol.~13, no.~4, 1998.

\bibitem{garroussi2020matheuristic}
Z.~Garroussi, R.~Ellaia, E.-G. Talbi, and J.-Y.~L. Lucas, ``A matheuristic for
  a bi-objective demand-side optimization for cooperative smart homes,''
  \emph{Electrical Engineering}, vol. 102, no.~4, 2020.

\bibitem{chiam2019hierarchical}
Z.~Chiam, A.~Easwaran, D.~Mouquet, S.~Fazlollahi, and J.~Mill{\'a}s, ``A
  hierarchical framework for holistic optimization of the operations of
  district cooling systems,'' \emph{Applied Energy}, vol. 239, 2019.

\bibitem{balamurugan2008hybrid}
R.~Balamurugan and S.~Subramanian, ``Hybrid integer coded differential
  evolution,'' \emph{Energy Conversion and Management}, vol.~49, no.~4, 2008.

\bibitem{gonzalez2022hyper}
M.~Gonzalez, J.~L{\'o}pez-Esp{\'\i}n, J.~Aparicio, and E.-G. Talbi, ``A
  hyper-matheuristic approach for solving mixed integer linear optimization
  models in the context of data envelopment analysis,'' \emph{PeerJ Computer
  Science}, vol.~8, 2022.

\bibitem{ye2019pso}
Y.~Ye, H.~Jinhou, C.~Chen, X.~Jiarong, L.~Zuwei, L.~Xinggao, C.~Jinshui, and
  L.~Jiangang, ``A {PSO-LP} cooperative algorithm for mixed integer nonlinear
  programming,'' in \emph{12th Asian Control Conference (ASCC)}, 2019.

\bibitem{garroussi2017hybrid}
Z.~Garroussi, R.~Ellaia, E.-G. Talbi, and J.-Y. Lucas, ``Hybrid evolutionary
  algorithm for residential demand side management with a photovoltaic panel
  and a battery,'' in \emph{Int. Conf. on Control, Artificial Intelligence,
  Robotics \& Optimization}, 2017.

\bibitem{zheng2013cooperative}
Y.-J. Zheng and S.-Y. Chen, ``Cooperative particle swarm optimization for
  multiobjective transportation planning,'' \emph{Applied Intelligence},
  vol.~39, no.~1, 2013.

\bibitem{ma2018survey}
X.~M. et~al., ``A survey on cooperative co-evolutionary algorithms,''
  \emph{IEEE Transactions on Evolutionary Computation}, vol.~23, no.~3, 2018.

\bibitem{legillon2012cobra}
F.~Legillon, A.~Liefooghe, and E.-G. Talbi, ``Cobra: A cooperative
  coevolutionary algorithm for bi-level optimization,'' in \emph{{IEEE}
  Congress on Evolutionary Computation (CEC)}, 2012.

\bibitem{potter1994cooperative}
M.~Potter and K.~K.~Jong, ``A cooperative coevolutionary approach to function
  optimization,'' in \emph{Int. Conf. on Parallel Oroblem Solving from Nature
  (PPSN)}, 1994.

\bibitem{hiremath2012designing}
N.~Hiremath, S.~Sahu, and M.~Tiwari, ``Designing a multi echelon flexible
  logistics network using co-evolutionary immune {PSO} with penetrated
  hyper-mutation,'' in \emph{Applied Mechanics and Materials}, vol. 110, 2012.

\bibitem{omidvar2013cooperative}
M.~Omidvar, X.~Li, Y.~Mei, and X.~Yao, ``Cooperative co-evolution with
  differential grouping for large scale optimization,'' \emph{IEEE Transactions
  on Evolutionary Computation}, vol.~18, no.~3, 2013.

\bibitem{van2004cooperative}
F.~V. den Bergh and A.~Engelbrecht, ``A cooperative approach to {PSO},''
  \emph{IEEE Transactions on Evolutionary Computation}, vol.~8, no.~3, 2004.

\bibitem{mei2016competitive}
Y.~Mei, M.~Omidvar, X.~Li, and X.~Yao, ``A competitive divide-and-conquer
  algorithm for unconstrained large-scale black-box optimization,'' \emph{ACM
  Transactions on Mathematical Software}, vol.~42, no.~2, 2016.

\bibitem{strasser2016factored}
S.~Strasser, J.~Sheppard, N.~Fortier, and R.~Goodman, ``Factored evolutionary
  algorithms,'' \emph{IEEE Transactions on Evolutionary Computation}, vol.~21,
  no.~2, 2016.

\bibitem{vinko2008global}
T.~Vink{\'o} and D.~Izzo, ``Global optimisation heuristics and test problems
  for preliminary spacecraft trajectory design,'' \emph{Advanced Concepts Team,
  ESATR ACT-TNT-MAD-GOHTPPSTD}, 2008.

\bibitem{pan2020effective}
Q.-K. Pan, L.~Gao, and L.~Wang, ``An effective cooperative co-evolutionary
  algorithm for distributed flowshop group scheduling problems,'' \emph{{IEEE}
  Transactions on Cybernetics}, 2020.

\bibitem{zhang2013hybrid}
X.~Zhang, X.~Guan, I.~Hwang, and K.~Cai, ``A hybrid distributed-centralized
  conflict resolution approach for multi-aircraft based on cooperative
  co-evolutionary,'' \emph{Science China Information Sciences}, vol.~56,
  no.~12, 2013.

\bibitem{yuan2020co}
S.~Yuan, T.~Li, and B.~Wang, ``A co-evolutionary genetic algorithm for the
  two-machine flow shop group scheduling problem with job-related blocking and
  transportation times,'' \emph{Expert Systems with Applications}, vol. 152,
  2020.

\bibitem{shi2017reference}
M.~Shi and S.~Gao, ``Reference sharing: a new collaboration model for
  cooperative coevolution,'' \emph{Journal of Heuristics}, vol.~23, no.~1.

\bibitem{popovici2012coevolutionary}
E.~Popovici, A.~Bucci, R.~Wiegand, and E.~D. de~Jong, ``Coevolutionary
  principles,'' in \emph{Handbook of Natural Computing}, G.~Rozenberg,
  T.~B{\"{a}}ck, and J.~N. Kok, Eds.\hskip 1em plus 0.5em minus 0.4em\relax
  Springer, 2012.

\bibitem{yang2008large}
Z.~Yang, K.~Tang, and X.~Yao, ``Large scale evolutionary optimization using
  cooperative coevolution,'' \emph{Information sciences}, vol. 178, no.~15,
  2008.

\bibitem{wiegand2001empirical}
P.~Wiegand, W.~Liles, and K.~D. Jong, ``An empirical analysis of collaboration
  methods in cooperative coevolutionary algorithms,'' in \emph{Genetic and
  Evolutionary Computation Conf. (GECCO)}, vol. 2611, 2001.

\bibitem{de2016cooperative}
F.~de~Oliveira, R.~Enayatifar, H.~Sadaei, F.~Guimar{\~a}es, and J.-Y. Potvin,
  ``A cooperative coevolutionary algorithm for the multi-depot vehicle routing
  problem,'' \emph{Expert Systems with Applications}, vol.~43, 2016.

\bibitem{son2004hybrid}
Y.~Son and R.~Baldick, ``Hybrid coevolutionary programming for nash equilibrium
  search in games with local optima,'' \emph{IEEE Transactions on Evolutionary
  Computation}, vol.~8, no.~4, 2004.

\bibitem{glorieux2015improved}
E.~Glorieux, B.~Svensson, F.~Danielsson, and B.~Lennartson, ``Improved
  constructive cooperative coevolutionary differential evolution for
  large-scale optimisation,'' in \emph{IEEE Symposium Series on Computational
  Intelligence}, 2015.

\bibitem{panait2006selecting}
L.~Panait and S.~Luke, ``Selecting informative actions improves cooperative
  multiagent learning,'' in \emph{5th Int. Joint Conf. on Autonomous Agents and
  Multiagent Systems}, 2006.

\bibitem{bucci2005identifying}
A.~Bucci and J.~Pollack, ``On identifying global optima in cooperative
  coevolution,'' in \emph{7th Annual Conference on Genetic and Evolutionary
  Computation}, 2005.

\bibitem{nguyen2007analysis}
M.~Nguyen, H.~Abbass, and R.~McKay, ``Analysis of ccme: Coevolutionary
  dynamics, automatic problem decomposition, and regularization,'' \emph{IEEE
  Transactions on Systems, Man, and Cybernetics, Part C (Applications and
  Reviews)}, vol.~38, no.~1, 2007.

\bibitem{song2020variable}
X.-F. Song, Y.~Zhang, Y.-N. Guo, X.-Y. Sun, and Y.-L. Wang, ``Variable-size
  cooperative coevolutionary particle swarm optimization for feature selection
  on high-dimensional data,'' \emph{IEEE Transactions on Evolutionary
  Computation}, vol.~24, no.~5, 2020.

\bibitem{abdelkhalik2012dynamic}
O.~Abdelkhalik and A.~Gad, ``Dynamic-size multiple populations genetic
  algorithm for multigravity-assist trajectory optimization,'' \emph{Journal of
  Guidance, Control and Dynamics}, vol.~35, no.~2, 2012.

\bibitem{hansen2008multilevel}
L.~Hansen and P.~Horst, ``Multilevel optimization in aircraft structural design
  evaluation,'' \emph{Computers \& structures}, vol.~86, no. 1-2, 2008.

\bibitem{sobieszczanski1997multidisciplinary}
J.~Sobieszczanski-Sobieski and R.~Haftka, ``Multidisciplinary aerospace design
  optimization: survey of recent developments,'' \emph{Structural
  optimization}, vol.~14, no.~1, 1997.

\bibitem{bandyopadhyay2001pixel}
S.~Bandyopadhyay and S.~Pal, ``Pixel classification using variable string
  genetic algorithms with chromosome differentiation,'' \emph{{IEEE}
  Transactions on Geoscience and Remote Sensing}, vol.~39, no.~2, 2001.

\bibitem{maulik2009modified}
U.~Maulik and I.~Saha, ``Modified differential evolution based fuzzy clustering
  for pixel classification in remote sensing imagery,'' \emph{Pattern
  Recognition}, vol.~42, no.~9, 2009.

\bibitem{falkenauer1994new}
E.~Falkenauer, ``A new representation and operators for {GA} applied to
  grouping problems,'' \emph{Evolutionary computation}, vol.~2, no.~2, 1994.

\bibitem{costa2018hierarchical}
V.~Costa and C.~Rodrigues, ``Hierarchical ant colony for simultaneous
  classifier selection and hyperparameter optimization,'' in \emph{{IEEE}
  Congress on Evolutionary Computation (CEC)}, 2018.

\bibitem{frank2016evolutionary}
C.~Frank, R.~Marlier, O.~Pinon-Fischer, and D.~Mavris, ``An evolutionary
  multi-architecture multi-objective optimization algorithm for design space
  exploration,'' in \emph{57th Structures, Structural Dynamics, and Materials
  Conf.}, 2016.

\bibitem{zebulum2000variable}
R.~Zebulum, M.~Vellasco, and M.~Pacheco, ``Variable length representation in
  evolutionary electronics,'' \emph{Evolutionary Computation}, vol.~8, no.~1,
  2000.

\bibitem{gad2011hidden}
A.~Gad and O.~Abdelkhalik, ``Hidden genes genetic algorithm for
  multi-gravity-assist trajectories optimization,'' \emph{Journal of Spacecraft
  and Rockets}, vol.~48, no.~4, 2011.

\bibitem{abdelkhalik2013hidden}
O.~Abdelkhalik, ``Hidden genes genetic optimization for variable-size design
  space problems,'' \emph{Journal of Optimization Theory and Applications},
  vol. 156, no.~2, 2013.

\bibitem{gamot2023}
J.~Gamot, M.~Balesdent, A.~Tremolet, R.~Wuilbercq, N.~Melab, and E.-G. Talbi,
  ``Hidden-variables {GA} for variable-size design space optimal layout
  problems with application to aerospace vehicles,'' \emph{Engineering
  Applications of Artificial Intelligence}, vol. 121, 2023.

\bibitem{abdelkhalik2013autonomous}
O.~Abdelkhalik, ``Autonomous planning of multigravity-assist trajectories with
  deep space maneuvers using a differential evolution approach,''
  \emph{International Journal of Aerospace Engineering}, vol. 2013, 2013.

\bibitem{chen2015reconfiguration}
Y.~C. et~al., ``Reconfiguration of satellite orbit for cooperative observation
  using variable-size multi-objective differential evolution,'' \emph{European
  Journal of Operational Research}, vol. 242, no.~1, 2015.

\bibitem{mukhopadhyay2014identifying}
A.~Mukhopadhyay and M.~Mandal, ``Identifying non-redundant gene markers from
  microarray data,'' \emph{IEEE/ACM transactions on computational biology and
  bioinformatics}, vol.~11, no.~6, 2014.

\bibitem{darani2018space}
S.~Darani and O.~Abdelkhalik, ``Space trajectory optimization using hidden
  genes genetic algorithms,'' \emph{Journal of Spacecraft and Rockets},
  vol.~55, no.~3, 2018.

\bibitem{rothlauf2006representations}
F.~Rothlauf, ``Representations for genetic and evolutionary algorithms,'' in
  \emph{Representations for Genetic and Evolutionary Algorithms}.\hskip 1em
  plus 0.5em minus 0.4em\relax Springer, 2006.

\bibitem{gao2021adaptive}
G.~Gao, Y.~Mei, Y.-H. Jia, W.~Browne, and B.~Xin, ``Adaptive coordination ant
  colony optimization for multipoint dynamic aggregation,'' \emph{{IEEE}
  Transactions on Cybernetics}, 2021.

\bibitem{gao2021hybrid}
G.~Gao, B.~Xin, Y.~Mei, S.~Ding, and J.~Li, ``A hybrid decomposition-based
  multi-objective evolutionary algorithm for the multi-point dynamic
  aggregation problem,'' \emph{arXiv preprint arXiv:2105.04934}, 2021.

\bibitem{reuter1997heterogeneous}
C.~Reuter, M.~Schwiegershausen, and P.~Pirsch, ``Heterogeneous multiprocessor
  scheduling and allocation using evolutionary algorithms,'' in \emph{IEEE Int.
  Conf. on Application-Specific Systems, Architectures and Processors}, 1997.

\bibitem{wang2018hybrid}
B.~Wang, Y.~Sun, B.~Xue, and M.~Zhang, ``A hybrid differential evolution
  approach to designing deep convolutional neural networks for image
  classification,'' in \emph{Australasian Joint Conference on Artificial
  Intelligence}, 2018.

\bibitem{gao2020memetic}
G.~Gao, Y.~Mei, B.~Xin, Y.-H. Jia, and W.~Browne, ``A memetic algorithm for the
  task allocation problem on multi-robot multi-point dynamic aggregation
  missions,'' in \emph{{IEEE} Congress on Evolutionary Computation (CEC)},
  2020.

\bibitem{merlevede2019homology}
A.~Merlevede, H.~{\AA}hl, and C.~Troein, ``Homology and linkage in crossover
  for linear genomes of variable length,'' \emph{Plos one}, vol.~14, no.~1,
  2019.

\bibitem{ryerkerk2019survey}
M.~Ryerkerk, R.~Averill, K.~Deb, and E.~Goodman, ``A survey of evolutionary
  algorithms using metameric representations,'' \emph{Genetic Programming and
  Evolvable Machines}, vol.~20, no.~4, 2019.

\bibitem{marek2022another}
M.~Marek and P.~Kadlec, ``Another evolution of generalized differential
  evolution: variable number of dimensions,'' \emph{Engineering Optimization},
  vol.~54, no.~1, 2022.

\bibitem{kiranyaz2009fractional}
S.~Kiranyaz, T.~Ince, A.~Yildirim, and M.~Gabbouj, ``Fractional {PSO} in
  multidimensional search space,'' \emph{IEEE Transactions on Systems, Man, and
  Cybernetics, Part B (Cybernetics)}, vol.~40, no.~2.

\bibitem{neumann2012targeted}
G.~Neumann and D.~Cairns, ``Targeted eda adapted for a routing problem with
  variable length chromosomes,'' in \emph{{IEEE} Congress on Evolutionary
  Computation}, 2012.

\bibitem{dwivedi2018learning}
P.~Dwivedi, V.~Kant, and K.~Bharadwaj, ``Learning path recommendation based on
  modified variable length genetic algorithm,'' \emph{Education and information
  technologies}, vol.~23, no.~2, 2018.

\bibitem{ryerkerk2012optimization}
M.~Ryerkerk, R.~Averill, K.~Deb, and E.~Goodman, ``Optimization for
  variable-size problems using genetic algorithms,'' in \emph{12th AIAA
  Aviation Technology, Integration, and Operations (ATIO) Conf.}

\bibitem{hutt2007synapsing}
B.~Hutt and K.~Warwick, ``Synapsing variable-length crossover: Meaningful
  crossover for variable-length genomes,'' \emph{{IEEE} transactions on
  Evolutionary Computation}, vol.~11, no.~1, 2007.

\bibitem{goldberg1989messy}
D.~Goldberg, B.~Korb, and K.~Deb, ``Messy genetic algorithms: Motivation,
  analysis, and first results,'' \emph{Complex systems}, vol.~3, no.~5, 1989.

\bibitem{harvey1992saga}
I.~Harvey, R.~Manner, and B.~Manderick, ``The saga cross: the mechanics of
  crossover for variable-length genetic algorithms,'' \emph{Parallel Problem
  Solving from Nature}.

\bibitem{burke1998putting}
D.~Burke, K.~D. Jong, J.~Grefenstette, C.~Ramsey, and A.~Wu, ``Putting more
  genetics into genetic algorithms,'' \emph{Evolutionary Computation}, vol.~6,
  no.~4, 1998.

\bibitem{dasgupta1992nonstationary}
D.~Dasgupta and D.~McGregor, ``Nonstationary function optimization using the
  structured genetic algorithm.'' in \emph{Parallel Problem Solving from Nature
  Conference (PPSN)}, vol.~2, 1992.

\bibitem{yan2009density}
Y.~Yan and L.~Osadciw, ``Density estimation using a new dimension adaptive
  {PSO} algorithm,'' \emph{Swarm Intelligence}, vol.~3, no.~4, 2009.

\bibitem{yangyang2004particle}
Z.~Y. et~al., ``{PSO} for base station placement in mobile communication,'' in
  \emph{IEEE InT. Conf. on Networking, Sensing and Control}, vol.~1, 2004.

\bibitem{kadlec2018particle}
P.~Kadlec and V.~{\v{S}}ed{\v{e}}nka, ``{PSO} for problems with variable number
  of dimensions,'' \emph{Engineering Optimization}, vol.~50, no.~3, 2018.

\bibitem{tran2018variable}
B.~Tran, B.~Xue, and M.~Zhang, ``Variable-length {PSO} for feature selection on
  high-dimensional classification,'' \emph{IEEE Transactions on Evolutionary
  Computation}, vol.~23, no.~3, 2018.

\bibitem{liang2006comprehensive}
J.~Liang, K.~Qin, P.~Suganthan, and S.~Baskar, ``Comprehensive learning
  particle swarm optimizer for global optimization of multimodal functions,''
  \emph{IEEE transactions on Evolutionary Computation}, vol.~10, no.~3, 2006.

\bibitem{talbi2006hierarchical}
E.-G. Talbi and H.~Meunier, ``Hierarchical parallel approach for gsm mobile
  network design,'' \emph{Journal of Parallel and Distributed Computing},
  vol.~66, no.~2, 2006.

\bibitem{liu2019coordinated}
W.-L. Liu, Y.-J. Gong, W.-N. Chen, Z.~Liu, H.~Wang, and J.~Zhang, ``Coordinated
  charging scheduling of electric vehicles: A mixed-variable differential
  evolution approach,'' \emph{IEEE Transactions on Intelligent Transportation
  Systems}, vol.~21, no.~12, 2019.

\bibitem{gentile2019structured}
L.~Gentile, C.~Greco, E.~Minisci, T.~Bartz-Beielstein, and M.~Vasile,
  ``Structured-chromosome ga optimisation for satellite tracking,'' in
  \emph{Genetic and Evolutionary Computation Conference (GECCO)}, 2019.

\bibitem{shan2006survey}
Y.~Shan, R.~McKay, D.~Essam, and H.~Abbass, ``A survey of probabilistic model
  building genetic programming,'' in \emph{Scalable optimization via
  probabilistic modeling}.\hskip 1em plus 0.5em minus 0.4em\relax Springer,
  2006.

\bibitem{koza1992programming}
J.~Koza, ``On the programming of computers by means of natural selection,''
  \emph{Genetic programming}, 1992.

\bibitem{isebor2009constrained}
O.~J. Isebor, ``Constrained production optimization with an emphasis on
  derivative-free methods,'' Ph.D. dissertation, Stanford University Stanford,
  CA, 2009.

\bibitem{kim2005variable}
Y.~Kim and O.~D. Weck, ``Variable chromosome length {GA} for progressive
  refinement in topology optimization,'' \emph{Structural and Multidisciplinary
  Optimization}, vol.~29, no.~6, 2005.

\bibitem{englander2012automated}
J.~Englander, B.~Conway, and T.~Williams, ``Automated mission planning via
  evolutionary algorithms,'' \emph{Journal of Guidance, Control, and Dynamics},
  vol.~35, no.~6, 2012.

\bibitem{chilan2013automated}
C.~Chilan and B.~Conway, ``Automated design of multiphase space missions using
  hybrid optimal control,'' \emph{Journal of Guidance, Control, and Dynamics},
  vol.~36, no.~5, 2013.

\bibitem{morar2021bayesian}
M.~Morar, \emph{Bayesian optimisation over mixed parameter spaces}.\hskip 1em
  plus 0.5em minus 0.4em\relax University of Manchester (United Kingdom), 2021.

\bibitem{manson2021mvmoo}
J.~Manson, T.~Chamberlain, and R.~Bourne, ``Mvmoo: Mixed variable
  multi-objective optimisation,'' \emph{Journal of Global Optimization},
  vol.~80, no.~4, 2021.

\bibitem{tanabe2020easy}
R.~Tanabe and H.~Ishibuchi, ``An easy-to-use real-world multi-objective
  optimization problem suite,'' \emph{Applied Soft Computing}, vol.~89, 2020.

\bibitem{li2019variable}
H.~Li, K.~Deb, and Q.~Zhang, ``Variable-length pareto optimization via
  decomposition-based evolutionary multiobjective algorithm,'' \emph{IEEE
  Transactions on Evolutionary Computation}, vol.~23, no.~6, 2019.

\bibitem{tang2016class}
Y.~Tang, J.-P. Richard, and C.~Smith, ``A class of algorithms for mixed-integer
  bilevel min-max optimization,'' \emph{Journal of Global Optimization},
  vol.~66, no.~2, 2016.

\bibitem{kleinert2021survey}
T.~Kleinert, M.~Labb{\'e}, I.~Ljubi{\'c}, and M.~Schmidt, ``A survey on
  mixed-integer programming techniques in bilevel optimization,'' \emph{EURO
  Journal on Computational Optimization}, vol.~9, 2021.

\bibitem{belotti2013mixed}
P.~Belotti, C.~Kirches, S.~Leyffer, J.~Linderoth, J.~Luedtke, and A.~Mahajan,
  ``Mixed-integer nonlinear optimization,'' \emph{Acta Numerica}, vol.~22,
  2013.

\bibitem{yao2011review}
W.~Yao, X.~Chen, W.~Luo, T.~Van, and J.~Guo, ``Review of uncertainty-based
  multidisciplinary design optimization methods for aerospace vehicles,''
  \emph{Progress in Aerospace Sciences}, vol.~47, no.~6, 2011.

\end{thebibliography}

\end{document}